\documentclass{article}

\usepackage{arxiv}

\usepackage[utf8]{inputenc} 
\usepackage[T1]{fontenc}    
\usepackage{hyperref}       
\usepackage{url}            
\usepackage{booktabs}       
\usepackage{amsfonts}       
\usepackage{nicefrac}       
\usepackage{microtype}      
\usepackage{graphicx}
\usepackage{doi}

\usepackage{amsmath,amssymb,amsfonts}
\usepackage{algorithmic}
\usepackage{textcomp}
\usepackage{caption}
\usepackage{comment}
\graphicspath{ {./images/} }

\title{A Survey on Automated Sarcasm Detection on Twitter}

\author{ \href{https://orcid.org/0000-0002-4131-7405}{\includegraphics[scale=0.06]{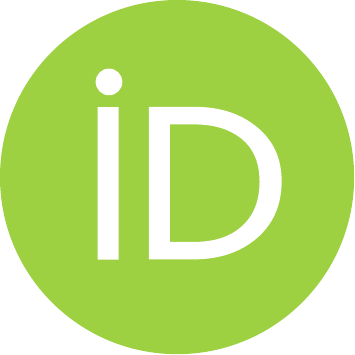}\hspace{1mm}Bleau Moores} \\
	Department of Computer Science\\
	Lakehead University\\
	Thunder Bay, ON P7B 5E1, Canada \\
	\texttt{bdmoores@lakeheadu.ca} \\
	\And
	\href{https://orcid.org/
0000-0002-9741-3463}{\includegraphics[scale=0.06]{orcid.pdf}\hspace{1mm}Vijay Mago} \\
	Department of Computer Science\\
	Lakehead University\\
	Thunder Bay, ON P7B 5E1, Canada \\
	\texttt{vmago@lakeheadu.ca} \\
}

\renewcommand{\shorttitle}{A Survey on Automated Sarcasm Detection on Twitter}

\hypersetup{
pdftitle={A Survey on Automated Sarcasm Detection on Twitter},
pdfsubject={cs.CL},
pdfauthor={Bleau Moores, Vijay Mago},
pdfkeywords={Deep Learning, Machine Learning, Natural Language Processing, Sarcasm Detection, Social Media, Twitter},
}

\begin{document}
\maketitle

\begin{abstract}
	Automatic sarcasm detection is a growing field in computer science.
Short text messages are increasingly used for communication, especially over social media platforms such as Twitter. Due to insufficient or missing context, unidentified sarcasm in these messages can invert the meaning of a statement, leading to confusion and communication failures.
This paper covers a variety of current methods used for sarcasm detection, including detection by context, posting history and machine learning models. Additionally, a shift towards deep learning methods is observable, likely due to the benefit of using a model with induced instead of discrete features combined with the innovation of transformers.
\end{abstract}


\section{Introduction} \label{introduction}
Social media has become a steadfast aspect of online life, with 97\% of online adolescent users regularly accessing social media sites \cite{Riehm2019}. Twitter, a social media site that serves as an online micro-blogging forum is ranked 10th for most internet traffic \cite{Fan2014}. Twitter is typically used to post about facts, opinions, events, ideas and humour, often accompanied by images, emotes, or videos \cite{Hee2018}. Additionally, Twitter has become a popular source of big data for analytics in marketing \cite{Park2016}, politics \cite{Small2011} \cite{Wang2012}, and stock predictions \cite{Oliveira2017} especially since Facebook, the most visited internet traffic site \cite{Fan2014}, eliminated access to their application programming interface's (API) in 2018 \cite{Freelon2018}. 
 Companies mine user generated data on politics \cite{Ott2016}, products \cite{Bonchi2011} \cite{Bouazizi2016} \cite{Qudar2021}, media \cite{Bouazizi2016}, and people \cite{Ghosh2017}. This mined data is explored for insights utilized in marketing and customer service, using sentiment analysis\cite{nguyen2015} \cite{yu2013} \cite{Xu2017} \cite{felix2017}.
 
 Sentiment analysis is performed by using a variety of Natural Language Processing (NLP) techniques and algorithms to determine whether a given string of text has a positive or negative sentiment attached to it and thus allows one to derive a sentiment about the subject of the text \cite{Cambria2017}. 
 Twitter data analysis is challenging as it is unstructured. However, the Twitter API provides an easily accessible means to collect large amounts of data from public posts \cite{Kim2020} \cite{Blank2017}. This data is generally succinct due to Twitter's character limit and often semi-accurately labelled due to user self-labelling via hashtags \cite{Small2011} \cite{qudar2020tweetbert}. Additionally, Twitter is used for building corpora via text mining, which further cements its utility in data science \cite{Joshi2017} \cite{Kovaz2013}.

For all these benefits of using Twitter data, there is still an ogre in the cupboard that must be addressed. Sarcasm is a form of irony or satire that inverts the sentiment of a post or remark, often in the form of a positive surface sentiment with a negative intended sentiment \cite{Kunneman2015} \cite{Riloff2013}. While common, as it is used for both humour and criticism over social media \cite{Bouazizi2016} \cite{Ghosh2017}, sarcasm is often missed by the reader, or listener \cite{Bouazizi2016}. This missed sarcasm leads to confusion as the author's or speaker's words are seen to support the surface sentiment instead of inverting it. When using data derived from social media networks, the prevalence of sarcasm introduces error in sentiment analysis and opinion mining \cite{Bouazizi2016}.

In verbal dialogue, sarcasm is often accompanied by context in the form of tone,  inflection or laughter \cite{Muresan2016}, facial expression and shared sentiments or beliefs \cite{Ghosh2016}. Even then, it can often go undetected as the listener requires knowledge that verifies or invalidates the surface intent of the statement \cite{Joshi2017}. For example, in the context of a baseball game, "{\it Way to go, Slugger}" has a very different meaning if directed at a player that hit a home run versus a player that struck out \cite{Riloff2013}. If the player had hit a home run, it would be a positive intended and surface sentiment. However, if the player had struck out, it would be a derisive or scornful intended sentiment conflicting with its positive surface sentiment. In both examples, the tone and context allow the listener to determine if it is sarcastic or not.

In microblog posts like Twitter, those nuances are often absent. This poses a problem for sentiment analysis algorithms. Some Twitter users compensate for the lack of tone and context by making use of hashtags like \#not, \#sarcasm, \#sarcastic, and \#irony \cite{Kunneman2015}. Other forms of context come from emotes like emoji (winking face, tongue out winking face, smiling face) or emoticons (:p, ;), :-), ;P ) \cite{Filik2016} to aid the reader in identifying sarcastic posts. Even using self-labelling techniques like hashtags or emotes, when using sarcasm, the writer of a Tweet assumes an element of risk that readers will not identify the sarcasm and thus will take the post literally. Often the reader assumes that risk with the trade-off of reducing emotional weight to the post, as the weight of the planned intent will be reduced somewhat by the paired opposing polarity surface intent \cite{Filik2016}.

\begin{figure}[h]
\includegraphics[width=14cm]{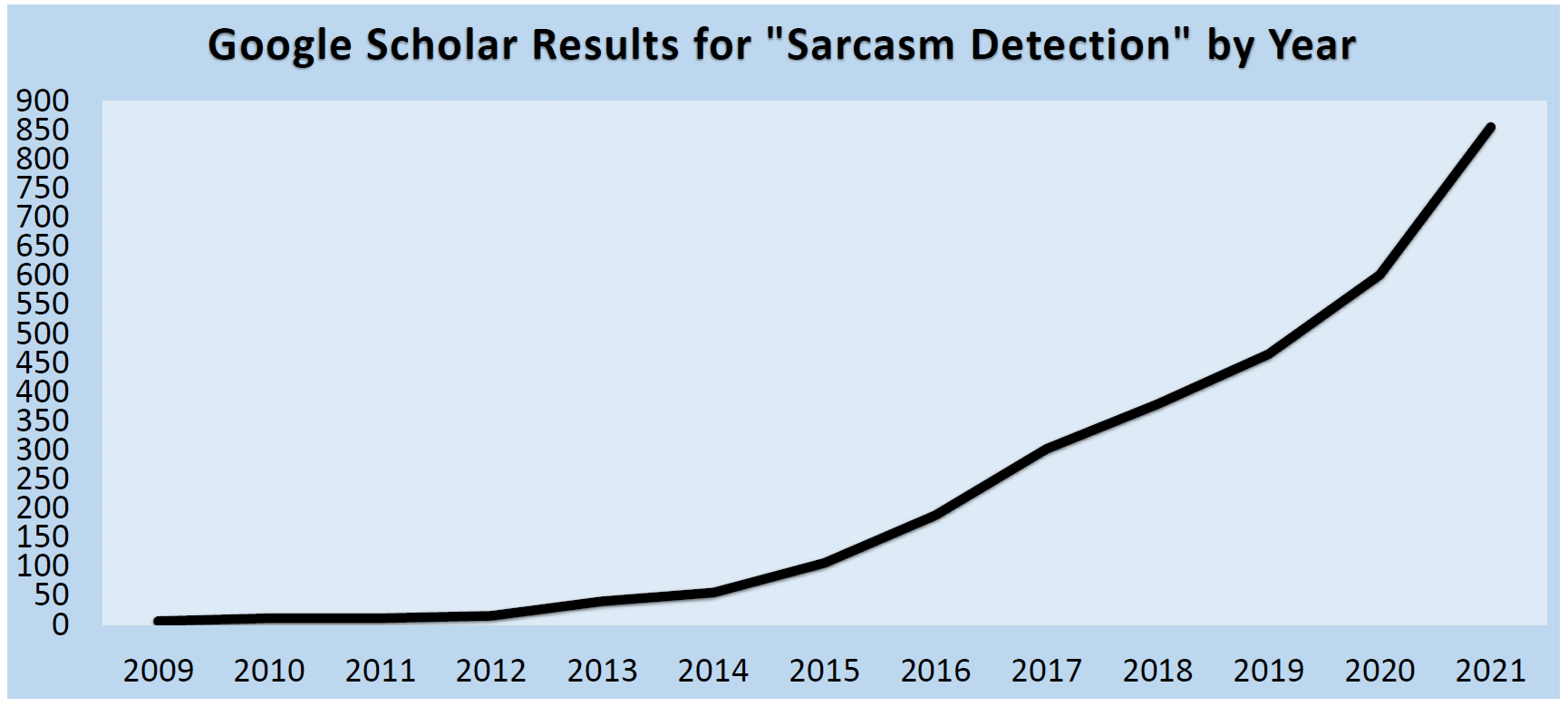}
\centering
\caption{Google Scholar Search Results for "Sarcasm Detection" by Year}
\label{GoogleScholarResults}
\end{figure}

Sadly, author-tagged Tweets, while helpful, are not standardized. Individual emotes have a variety of uses beyond sarcasm, and hashtags may be used ironically or inaccurately. Thus a reliable method of automatic sarcasm detection is still a challenge, in both design and implementation, to be overcome \cite{Kunneman2015}. Even in research environments with human-annotated Tweets, sarcasm detection struggles due to the innate subjectivity of the classification \cite{Kovaz2013}, which can lead to inaccurate labelling. An automated method for reliably and accurately flagging text as sarcastic is necessary for predicting accurate text sentiment \cite{Joshi2017}. Undetected sarcasm, by its nature, will create errors in sentiment analysis due to sarcastic statements being misinterpreted as literal statements \cite{Riloff2013}. Sarcasm detection is considered a challenging problem as it may rely on a hybrid of both lexical and context detection methods \cite{Muresan2016}.While automated sarcasm detection is still a new area of research \cite{Muresan2016}, Figure \ref{GoogleScholarResults} highlights the rapid growth in research on this subject. 

Undetected sarcasm introduces error during sentiment analysis as NLP models will interact with the surface sentiment instead of the intended sentiment \cite{Riloff2013}. It is imperative to develop accurate and efficient sarcasm detection algorithms is to mitigate these errors.

Two notable previous surveys into the topic exist: "Automatic Sarcasm Detection: A Survey" provides an in-depth and detailed investigation into approaches using rule-based, deep learning and statistical methodologies \cite{Joshi2017} while "Literature Survey of Sarcasm Detection" provides a quick overview of current detection approaches \cite{Chaudhari2018}. Neither survey provides a methodology for article selection, allowing for the inclusion of low-quality articles and studies. Additionally, both studies were published in 2017 and thus lack insight into more recent approaches. A newer survey of note is "Sarcasm detection using machine learning algorithms in Twitter: A systematic review," \cite{Sarsam2020} which provides a comprehensive overview of older machine learning models. Unfortunately, it does not incorporate any of the newer models using transformers.
Other than these notable surveys, a handful of less comprehensive surveys \cite{aboobaker2020} \cite{Chaudhari2017} \cite{Jansi2018} \cite{Haripriya2017} \cite{Kumar2020-2} \cite{Sentamilselvan2021} have been published in less reputable venues or are awaiting publication. These surveys offer very little insight into sarcasm detection methodologies. Instead, they offer a brief overview, usually 1-2 paragraphs on older approaches. This survey will be restricted to only including papers and articles from high-quality venues, as detailed in the methodology section.

The remainder of this paper will be as follows: Section \ref{Twitter} will discuss Twitter and methods of mining Tweets. Discussing sarcasm in depth will be the focus of Section \ref{Aspects of Sarcasm}, including building a definition for sarcasm and highlighting its characteristics, methods of usage and identification, and available resources for sarcasm datasets. This will be followed by a review into modern methods of modelling and classification used for sarcasm detection, in Section \ref{Modern Modelling Techniques used in Sarcasm Detection}. Section \ref{Summation of Modern Sarcasm Detection Models} will be devoted to an analysis of current automatic sarcasm detection methods, including the evaluation metrics and data sources used. Concluding remarks bring this survey to a close in Section \ref{Concluding Remarks}. At the end of the article, an appendix is included with the features of a Tweet for those unfamiliar with the platform.

\section{Twitter} \label{Twitter}
 Twitter, is a microblogging social media network initially launched in June 2006 and since then has experienced massive growth. It allows for the expression of opinions and facts, either open blogged or in response to other users \cite{Rajadesingan2015}. Common uses of Twitter include political discourse \cite{McCormick2017} \cite{Bolsover2019} \cite{Pond2019}, socialization \cite{Lara-Cabrera2019} \cite{ARORA201986} \cite{Cheng2021}, consumer engagement \cite{Wadhwa1866} \cite{read_robertson_mcquilken_ferdous_2019} \cite{oviedo2017}, and real time news sharing \cite{Kwak2010} \cite{Kalsnes2018} \cite{ORELLANARODRIGUEZ201874}.

\subsection{ Twitter Data Collection}

For any project involving Twitter, access to a large dataset of Tweets is required. Fortunately, resources are available for building a Twitter dataset through Twitter developer APIs. These options can provide a dataset that is either a general, untargeted sampling of what is available or targeted samples based on hashtags, usernames, and location. Due to the Twitter developer policy, when sharing datasets, only Tweet IDs may be shared, not the Tweets themselves\footnote{https://developer.twitter.com/en/developer-terms/agreement-and-policy}. This policy may cause issues in future studies using the same dataset. Some Tweets may become unavailable over time due to changes in availability, possibly due to user bans, Tweet deletions or author privacy settings.

Access to the API is provided by registering an account. As of the end of 2021, there are three account levels: Essential, Elevated and Academic Research. Essential accounts are free and allow access to a single project with a 500,00/month Tweet limit. Elevated accounts are enterprise-level, providing access to up to 2,000,000 Tweets/Month and require an approved developer account application\footnote{https://developer.twitter.com/en/support/twitter-api/developer-account\#faq-developer-account}. Academic Research accounts allow access to up to 10,000,000 Tweets/month. However, Academic Research accounts are restricted to graduate students and academic researchers with clearly defined objectives or use cases. Unlike Essential and Elevated accounts, Academic Research access accounts may not be used for commercial purposes\footnote{https://developer.twitter.com/en/products/twitter-api/academic-research}.

Common methods of accessing Tweets for research include:
\begin{itemize}
\item \textbf {Real-time Tweets API- } As of the end of 2021, Twitter currently provides two real-time Tweet streaming API, PowerTrack\footnote{https://developer.twitter.com/en/docs/twitter-api/enterprise/powertrack-api/overview} and Filtered stream endpoint\footnote{https://developer.twitter.com/en/docs/twitter-api/tweets/filtered-stream/introduction}. PowerTrack is an enterprise-grade real-time API available to Elevated and Academic Research accounts. Powertrack allows for up to 250,000 filter rules on a stream. Enterprise access allows for all standard filtering parameters and premium operators, such as keyword, emoji, exact phrase matching, hashtag and user filters. Filtered stream is a REST API endpoint accessible to all users, allowing users to set rules for Tweet filtering. The Rule limit is dependent on the account access, with Essential accounts having a 5 rule limit, Elevated a 25 rule limit and Academic Research a 1,000 rule limit. These rules allow for an expanded list of stand-alone operators, including; those previously mentioned under standard and enterprise access, URL, to, from, context, place and entity. Entity operators must be associated with a stand-alone operator and allow for filtering of Tweets by the inclusion of images, videos, mentions, links, reTweets, etc., that are included in the content of a Tweet.

 \item \textbf {Twitter Historical API-} Twitter provides access to both a Full-Archive search API and Historical PowerTrack Search API. Both provide access to all publicly available Tweets from March 2006 onward. The Full-Archive Search allows searching by term (effectively one rule), returning results matching the rule in no particular order. In contrast, the Historical PowerTrack allows up to 1000 rules and returns time-series data in 10 minute periods. As the Historical Powertrack API is job-based, a constant connection is not required as it is for the stream-based APIs \cite{Kim2020}.
 
 \item \textbf {Twitter Public API-} Twitter crawling makes use of the public API and allows for recursively mining a user or hashtag. This is useful for gathering historical or response Tweets for determining context.
 
 \item \textbf {TwitteR \cite{gentry} }TwitteR is an R package used to extract Tweets based on search parameters using the Twitter API, storing them in a database or .CSV file. Search parameters include language, time window start and end dates, geocode and ID (64-bit timestamps) windows. Be cautious, as Twitter extracts random Tweet samples during a time window, this can lead to duplicate Tweets for requests made during the same or overlapping windows, and data must be cleaned appropriately \cite{Ozturk2018}.
 
\item \textbf {Tweepy\footnote{https://www.tweepy.org/}-}Tweepy is a Python library used to access the Twitter API. Tweepy gives access to the entire Twitter catalogue of REST APIs and allows for both Tweet streaming and crawling. Tweepy also supports Tweet hydrating, which is the process of downloading a Tweet from its ID. In order to make use of the PowerTrack API, an enterprise-level login is required.
\end{itemize}

\section{Aspects of Sarcasm}\label{Aspects of Sarcasm}
\subsection{Defining Sarcasm}

Sarcasm is a form of irony \cite{Filik2016} \cite{Joshi2016} generally associated with an intention to mock or insult in a contemptuous or caustic manner using wit \cite{Rajadesingan2015} \cite{Riloff2013}, often for the purposes of criticizing a person or situation. However, it is also used as an expression of aggressive humour \cite{Rajadesingan2015}, and occasionally praise \cite{Filik2016}.
Sarcasm in its base form is a statement polarity inverter \cite{Bouazizi2016}, meaning that its surface or literal sentiment is the opposite of its intended sentiment \cite{Kunneman2015}. Thus a sarcastic statement means the opposite of what is stated \cite{Filik2016}. Additionally, the statement must be purposely constructed to have opposing surface and intended sentiments, thus being the deliberate intention of the speaker or writer. A statement misinterpreted by its audience is not sarcastic \cite{Joshi2017}.
Sarcasm is generally produced in 3 variants:
\begin{itemize}
 \item \textbf {A Positive Surface Sentiment With Negative Intended Sentiment-} This is the most common variant of sarcasm found on Twitter \cite{Riloff2013} and is typically used derisively to mock, ridicule or criticize \cite{Joshi2017} \cite{Rajadesingan2015} \cite{Riloff2013}. Examples would include "Mondays are the best!!!!" and "I LOOOVE your new haircut". 
 
 \item \textbf {A Negative Surface Sentiment With Positive Intended Sentiment- }Joshi \cite{Joshi2017} excludes this from sarcasm, instead categorizing it as humblebragging as it lacks a negative intended sentiment, as in the statement "I hate having a job, where I'm well paid to perform work I enjoy" \cite{Joshi2017}. However, most other studies include it as an aspect of sarcasm that can be used for purposes of praise, as in "Wow, your horrible at this game" after losing a match to an opponent \cite{Filik2016}.

 \item \textbf {A Neutral Surface Sentiment With Negative Intended Sentiment -} Sarcasm in the form of neutral surface sentiment with negative intended sentiment is generally used as a response during a dialogue, as in the case "... and I am the king of Egypt" \cite{Joshi2017}. This variant is expressly excluded from automatic detection in some studies due to the relative ease in detection by the audience compared to the polarity switched Tweets \cite{Muresan2016}.

\end{itemize} 
For the purposes of this research, the emphasis of sarcasm detection is on detecting deliberate polarity inversion between the surface and intended sentiments. While the positive surface sentiment with negative intended sentiment variant is the most common variant and some studies exclusively identify this variant, creating a unified way of detecting both of the opposing sentiment variants would be more efficient. After all, determining the intended sentiment is the goal, regardless of polarity, as using the intended sentiment is essential to reduce error in sentiment analysis and opinion mining \cite{Maynard2014}. Thus, sarcasm shall be defined as a statement deliberately composed to have an opposing surface or literal sentiment to its intended sentiment used intentionally to criticise, praise or evoke humour \cite{Bouazizi2016}.

With sarcasm defined, the next step is to investigate methods of identifying sarcasm that would apply to Twitter. From a linguistic approach, text-based sarcasm generally contains one or more of the following features: propositional, embedded, illocutionary, like prefixed, echo mention theory and dropped negation \cite{Joshi2017}. While these features do not identify sarcastic statements on their own, they may be used to identify statements that may be sarcastic. A brief discussion of these features follows:

\begin{itemize}
 \item \textbf {Propositional Sarcasm Features- }Propositional sarcasm occurs when the author puts forth a statement that, without context, will not be detected as sarcastic \cite{Joshi2017}. Devoid of context, for example, "That's a great idea" is challenging to identify as sarcastic or sincere. In Twitter, additional context cues in the form of emojis or hashtags may be required to provide context if the author intended sarcasm or not.
 \item \textbf {Illocutionary Sarcasm Features- }The "Way to go slugger" example from earlier is an example of illocutionary sarcasm. This type of sarcasm is when a sincere statement is conveyed alongside a non-textual context cue, such as a sigh, eye movement (rolled eyes, glare), smirking, or suprasegmental linguistic features (tone, inflection, stress) \cite{Matsui2016}. On Twitter, the text alone without a context clue (emoji, hashtag) is challenging to identify as sarcastic.
 \item \textbf {Like-Prefixed Sarcasm Features- }Like-prefix sarcasm is similar to propositional sarcasm but easier to detect. It is presented in the form of a declarative sentence and prefaces the sarcastic remark with a sarcastic signaller such as "like" or "sure." The sarcastic signaller inverts the statement being made \cite{Joshi2017}. In the example "like you care", the intended sentiment is that the subject does not care.
 \item \textbf {Echo Mention Theory Features- }Echo mention theory uses situations that remind the audience of past situations but with sarcastic clues. These often make use of hyperbole or sarcastic patterns of positive sentiment towards a negative situation \cite{Joshi2017} \cite{Kreuz1989}. "I love how when I put three pairs of socks in the dryer, and I pull out five mismatched individual socks." This reminds the audience of common jokes or life experiences with missing socks after laundry but in a hyperbolic manner.
 \item \textbf {Dropped Negation Sarcasm Features- }In dropped negation, a statement is made without a negation expression \cite{Joshi2017}. This often takes the form of a positive sentiment to a negative situation "I love being told my ideas are worthless." While self-labelling and context clues can assist here, a pattern-based approach may detect this better.
 \item \textbf {Hyperbole- }Hyperbole or over-exaggeration for emphasis may be a marker for sarcasm. An author's statement of "Wow, you are so good at that" could be either a sincere compliment or a sarcastic critique depending on the context. On its own, hyperbole is insufficient to determine if a statement is sarcastic, but it can be used to flag statements for further investigation \cite{Joshi2016}.

 \item \textbf {Pragmatic Feature- }Pragmatic features include additional aspects that accompany the text like emotes, or laughing features ("LOL," "hahaha," "hehe"), all word capitalization ("SERIOUSLY") or excessive punctuation marks ("!!!!") that perform as a lexical sarcasm marker \cite{Joshi2015}.
\end{itemize}

\subsection{Sarcasm Self Identifying Features on Twitter}
Often, but not always, Tweets will be author annotated as sarcastic. The two most common forms of author sarcasm annotation are hashtags and emotes, which deserve further elaboration.

\noindent \textbf{\#Hashtags:} Authors often use hashtags to label a Tweet as sarcastic. Typical sarcasm labelling hashtags are \#sarcasm, \#sarcastic and \#not. While \#sarcasm and \#sarcastic are evident in their intent \#not is a meta-communication marker \cite{Kunneman2015} that in a sense is adding back in a dropped negation. Other hashtags that may imply sarcasm are \#irony and \#cynicism as these are related linguistic mechanisms \cite{Kunneman2015}. Care must be taken when using hashtags as a labelling mechanism, as they may lead to errors in annotation. As hashtag use is not standardized, a situation that may lead to error is a hashtag not being used to label a post as sarcastic but to comment on sarcasm or a sarcasm property. "I wish there was a \#sarcasm font" and "Forgive me, I'm just feeling \#sarcastic today" are examples of Tweets that include sarcasm hashtags but are intended as sincere, not sarcastic. Deprived of context, a reader may find these sarcastic even if they are not intended as such. Another issue is incorrectly labelled posts; some authors may not label with hashtags; thus, a sarcastic post could be labelled as sincere. Additionally, some authors do not fully understand sarcasm and may incorrectly self-label with a hashtag. Hashtags have also been used for determining non-sarcastic Tweets by using hashtags containing either positive or negative sentiment words such as \#excited and \#annoyed \cite{Ghosh2018} \cite{Muresan2016}.
 
\noindent \textbf{Emotes:}
Emotes are a category that includes both emojis and emoticons. Emojis are small cartoon pictures, such as smiling faces, angry faces, and winking faces. Similarly, emoticons are the predecessor of emoji and are created from ASCII symbols like :(, :), and ;P, which are still in use today. Both emoji and emoticons may help determine if a Tweet is sarcastic. If the author intends to sarcastically praise or criticize, they may use an emote to serve as a context cue to the true intent of the text. This use of emoticons is illustrated in the examples "I'm so happy you arrive 30 minutes late! :(" or "You won the tournament? But you're such a bad player :)." Other context clues to sarcasm can include a winking ;) or tongue face emote :p \cite{Filik2016}. However, like hashtags, these are not standardized features, and to add additional ambiguity, emojis and emoticons may be used for a wide variety of uses beyond labelling sarcasm.

\subsection{Data Sources}
Datasets are required to train, test and validate models, and finding large, accurate datasets can be a challenge. Fortunately, there are online resources available to assist in curating a dataset. The rest of this section provides information about some common sarcasm-annotated dataset resources.
\noindent \textbf{Twitter Crawling - Author Annotated:}
Author annotated Twitter crawling occurs when the dataset is collected via Twitter developer APIs to build a Tweets database, then use Tweet annotation in the form of hashtags or emotes to label as sarcastic or non-sarcastic/sincere. Sarcastic Tweets are generally labelled via the hashtags \#sarcasm, \#sarcastic, \#not and \#irony \cite{Joshi2017} \cite{Hee2018}. In contrast, non-sarcastic Tweets are generally labelled with hashtags involving positive and or negative sentiments like \#angry, \#happy, \#blessed, and \#frustrated. Additional cleaning must be done here to reduce error. Initial data cleaning steps include removing duplicates, reTweets, and quotes \cite{Ghosh2016} \cite{Ghosh2017}\cite{Son2019}. Other filtering steps may include excluding short Tweets (less than three words), Tweets that only include URLs, Tweets not in the language of focus, and Tweets that have the labelling hashtags located somewhere other than the end of the message \cite{Ghosh2018}.

\noindent \textbf{The Internet Argument Corpus (IAC and IAC\textsubscript{v2}):}
IAC and IAC\textsubscript{v2}\footnote{https://nlds.soe.ucsc.edu/iac} are online public access corpora built on forum conversations concerning political and social topics. Each post contains the prior post in a conversation to provide context and is annotated in a variety of manners, including agreement and sarcastic \cite{Walker2012}. A subset of the IAC\textsubscript{v2} is the Sarcasm Corpus V2 \cite{slukin} which was derived by first flagging posts as sarcastic or not by using supervised pattern learners on IAC posts. Following the machine learning annotation, crowdsourced annotation was utilized to verify whether the post was sarcastic or sincere. This created a corpus of 4692 annotated posts, balanced between sarcastic and non-sarcastic, available for public use \cite{Oraby2016}. It is important to note that as IAC\textsubscript{v2} is annotated through crowdsourcing, it is labelled by the audience's perceived intent and not necessarily the author's intent \cite{Ghosh2018}.

\noindent \textbf{Self-Annotated Reddit Corpus (SARC):}
SARC\footnote{https://nlp.cs.princeton.edu/SARC/} is a collection of labelled sarcastic and non-sarcastic Reddit posts and comments released by Khodak, Saunshi, and Vodrahalli \cite{Ghosh2018}. Sarcasm labelling is performed by adding a trailing "/s" after the post's text. While sarcasm self-labelling is a common posting convention, a noise component is involved. Not all users abide by this convention, which leads to non-sarcastic labelled sarcastic posts. When checked against human-annotated labelling, a false negative rate of 2\% and a false positive rate of 1\% were determined to exist \cite{Khodak2019}.

\noindent \textbf{GoodReads:}
Goodreads\footnote{https://www.goodreads.com/} is an online social networking site that allows users to discuss and share information about books. Users post quotes, reviews, recommendations and can share opinions, theories and insights about books with like-minded individuals \cite{thelwall2016}. As many quotes are user annotated as sarcastic, they can be used to create a dataset for training, and testing sarcasm detection methods \cite{Joshi2016}.

\noindent \textbf{Google Books:}
Google Books\footnote{https://books.google.com/} is a massive repository of text drawn from full books and magazines from over the last 500 years \cite{Lin2012}. The repository contains over half a trillion words across a variety of languages, 361 billion in English, and is composed of text from over 5 million books \cite{Pechenick2015}. The large number of words from a variety of sources makes it very valuable for its use in corpus building for NLP tasks.

\noindent \textbf{SemEval 2015:}
SemEval is an annual research workshop designed to further advancement in semantic analysis and create high-quality annotated datasets for use in NLP. SemEval 2015 task 11 included a training dataset of 8,000 crowdsourced annotated English Tweets, of which 5,000 were labelled as sarcastic, and a testing dataset of an additional 4,000 annotated Tweets with 1,200 labelled as sarcastic \cite{ghosh-etal-2015-semeval}.

\section{Modern Modelling Techniques used in Sarcasm Detection} \label{Modern Modelling Techniques used in Sarcasm Detection}

Recent approaches to sarcasm detection involve extracting features from text (and sometimes earlier context text) and then processing them through a model to determine if the text is sarcastic or not \cite{Ghosh2018} \cite{Ren2018} \cite{Ren2016}. While earlier works often used regression analysis for classification \cite{Gonzalez-Ibanez2011} \cite{Kovaz2013} \cite{Rajadesingan2015}, there has been a  decline in its use in modern models for sarcasm detection. Instead, an increase in the use of more sophisticated machine learning and deep learning models have become more common in recent sarcasm detection methods \cite{kumar2020} \cite{liu2019} \cite{Ren2018} \cite{Zhang2016}. Machine learning algorithms such as support vector machines (SVM) \cite{Joshi2016}, neural networks (NN) \cite{Ghosh2016}, and random forest \cite{Bouazizi2016} were common in earlier models. In more recent approaches, deep learning models such as Convolutional Neural Networks (CNN) \cite{Poria2016}, and Long Short Term Memory (LSTM) \cite{Ghosh2018} have been commonly observed, often because of the benefits of automatic feature extraction\cite{Fisher2020}. This section of the article will explain some of the different machine learning and deep learning methodologies used for classification in sarcasm detection.


\subsection{Neural Networks}

Neural networks are machine learning algorithms loosely designed off the structure of a biological brain neuron \cite{MaassWolfgang1997Nosn}. The typical topology consists of an input layer, hidden layer(s) and output layer \cite{Orhan2011} \cite{Salama2002}. Each layer comprises interconnecting nodes or neurons, with each of the connections having a weight assigned to it. The input layer contains one neuron for every feature of the input data. Thus a data point for $X$, $Y$ and $Z$ coordinates (three features) would have three input neurons \cite{Salama2002}. The hidden layer(s) always have $n$ layers, and each layer consists of at least one neuron. The number and size of hidden layers vary between applications as having too few layers or nodes in a layer can lead to overfitting, but adding excessive complexity increases training time and resources \cite{Joshi2020}. A perceptron consists of a single hidden layer composed of a single neuron. In contrast, a single-layer perceptron (SLP) consists of a single hidden layer of multiple neurons, and a multi-layer perceptron (MLP) network consists of multiple hidden layers of at least one perceptron. In an MLP network, the neurons of the hidden layers contain a summation function for the products of all the input connections $x$ multiplied by the connections weight $w$ plus the neuron bias $b$ (if applicable). An activation function $f$, such as $sigmoid$, $sin$, $signum$ or $tanh$ \cite{Bianchini2014} \cite{Glorot2010} \cite{Yu2019}, then processes the result of the summation to provide an output $y$ as show in Equation \ref{nneq}. The output layer will either feed into different networks inputs or provide a usable output. In the case of classification networks, the outputs are often $softmax$, $signum$ or $tanh$. Individual connection weights are adjusted during model training via either feed-forward or backpropagation algorithms across numerous epochs \cite{Glorot2010} \cite{Hunter2012}.

\begin{equation}\label{nneq}
{\mathit{y_{j} = f\left ( \sum w_{ji}x{_{i}+b} \right )}}
\end{equation}

\subsection{Random Forest (RF)}

\begin{figure}[h!]
\includegraphics[width=7cm]{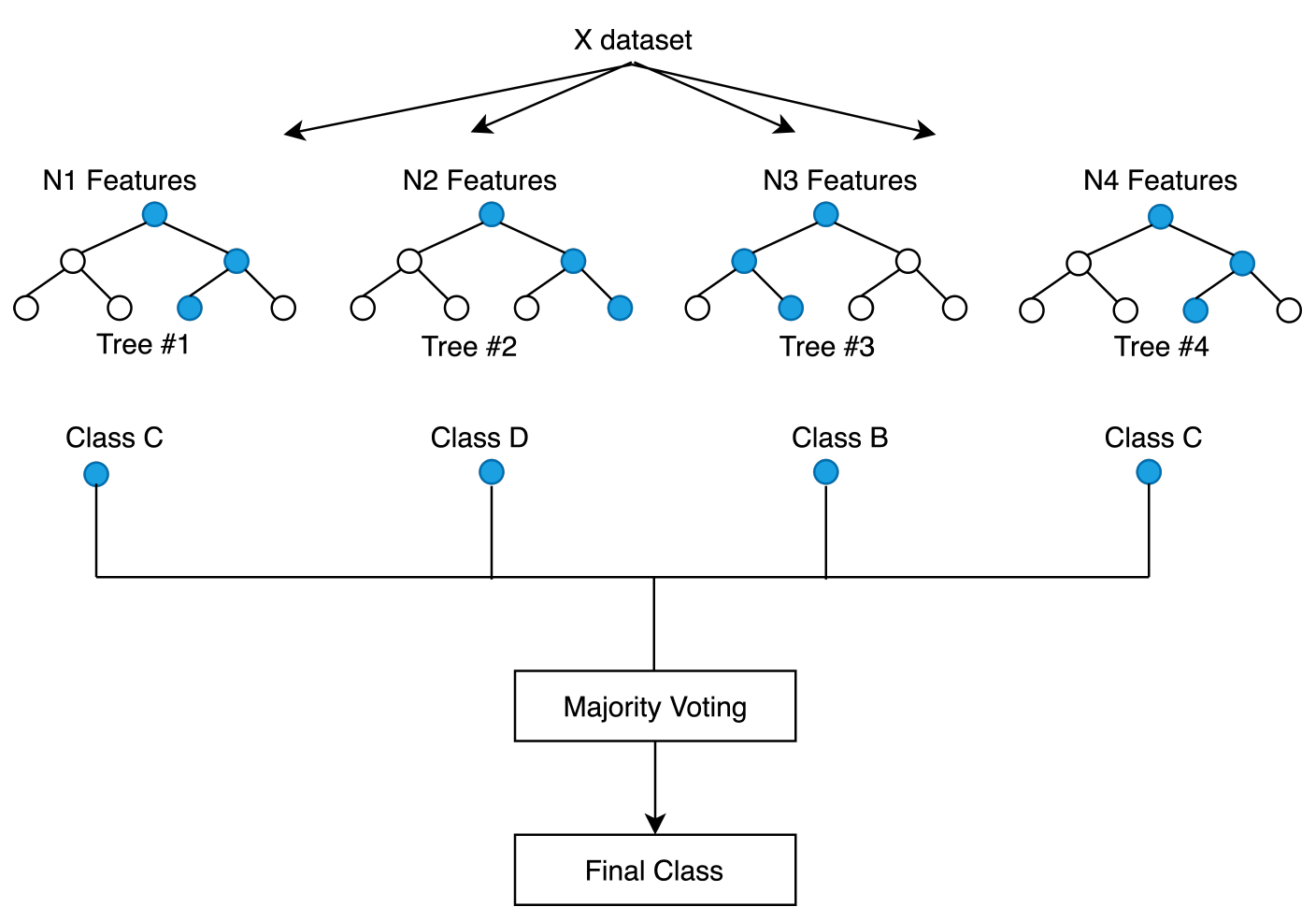}
\centering
\caption[Random Forest Architecture]{Random Forest Architecture }
\label{RF}
\end{figure}

Random Forest is a supervised machine learning algorithm where an ensemble of decision trees (the forest) are built with each tree being constructed using a random subset of training samples and variables \cite{Belgiu2016} \cite{Xie2019}. The trees are then merged into a single structure. Each individual tree provides a prediction of the data points class with a higher number of trees reducing output variance \cite{Couronne2018} \cite{Probst2018}. A set of hyperparameters defines the forest and the individual tree structure. Some typical hyperparameters are the number of observations for each tree, the number of samples contained in a node and the number of trees in the forest  \cite{Probst2019}. The class candidate that receives the most predictions or votes from the forest is considered the winning candidate and data point is categorized as such (Fig. \ref{RF}) \cite{Belgiu2016} \cite{Garg2021} \cite{Mishra2019} \cite{naghibi2017} \cite{pal2005}.


\subsection{Support Vector Machines (SVM)}

\begin{figure}[h!]
\includegraphics[width=8cm]{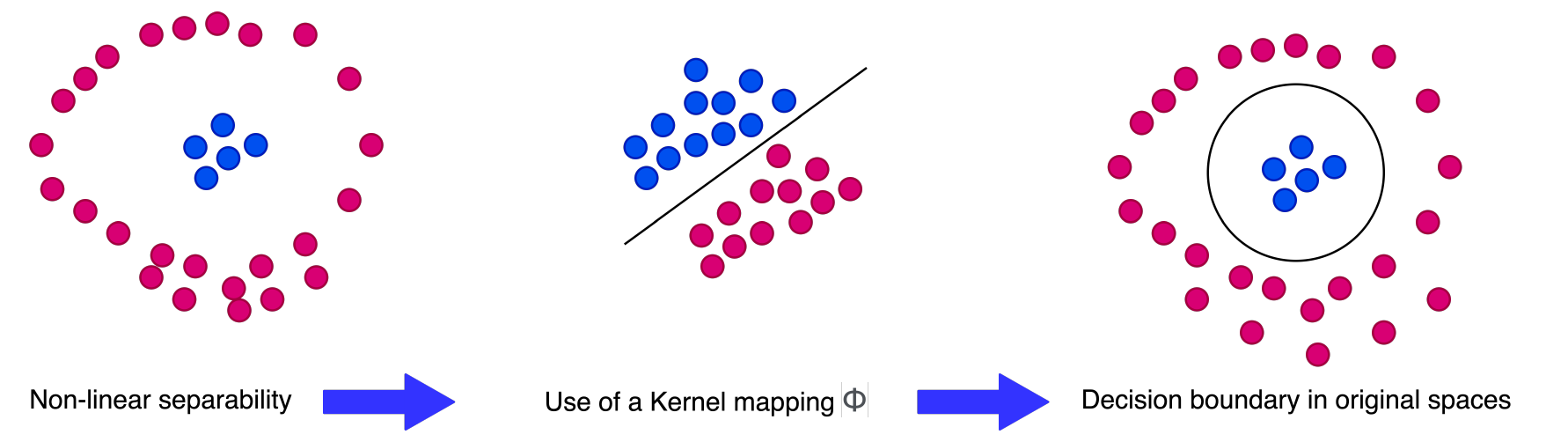}
\centering
\caption[SVM Mapping]{SVM Mapping }
\label{SVM}
\end{figure}

SVMs are a supervised machine learning algorithm used as a classifier \cite{tassone_yan_simpson_mendhe_mago_choudhury_2020} making use of the principle of structural risk minimization, which minimizes the upper bound on the generalization error \cite{Hsu2002} \cite{Osuna1997} \cite{Zendehboudi2018}. It takes the features of a non-linear separated dataset and maps it to a higher-dimensional space. In this higher-dimensional space, a hyperplane is determined, via non-linear mapping, that can classify the data points into two distinct classes \cite{CHRISTOPHERJ.C.BURGES1998} \cite{Osuna1997} \cite{Zendehboudi2018}. The higher dimensional space is then projected back to the original lower-dimensional space. The ideal hyperplane has the greatest margin between the two classes. The margin is defined as the sum of the distances from the hyperplane to the closest point of each class\cite{Osuna1997}. Determining the greatest margins is done using an application of a kernel function; some of the commonly used kernels can be seen in Table \ref{KernalFormulas}.

\begin{table*}[h]
\centering
\begin{tabular}{p{5cm} l}
\hline
Kernel                                   & Formula                                                                                                                                                                                                                                                                                                                                                                                                \\ \cline{1-2}\\
Polynomial kernel                        & 

$k\left ( x_{i},x_{j} \right ) = \left ( x_{i}\cdot x_{j}+1 \right )^{d} $\\

&  \\
Gaussian kernel                          & 

$k\left ( x,y \right ) = exp\left ( - \frac{\left \| x-y \right \|^{2}}{2\sigma ^{2}} \right )$\\

&  \\
Laplace RBF kernel                       & 
$k\left ( x,y \right ) = exp\left ( - \frac{\left \| x-y \right \|}{\sigma } \right )$\\

&  \\
Hyperbolic tangent kernel                & $k\left ( x_{i},x_{j} \right ) = tanh\left ( \kappa  x_{i}\cdot x_{j} +c\right )    $      \\                                                                                                                                                                                                   &  \\
Sigmoid kernel                           & $k\left ( x,y \right ) = tanh\left ( \alpha x^{\tau } y+c\right )$\\                                                                                                                                                                                              &  \\
Bessel function of the first kind kernel & $k\left ( x,y \right ) = \frac{J_{v+1}\left ( \sigma \left \| x-y \right \| \right )}{\left \| x-y \right \| ^{-n(v+1)}}$\\                                  &  \\

ANOVA radial basis kernel                & $k\left ( x,y \right ) = \sum_{k=1}^{n} exp\left ( -\sigma \left ( x^{k}-y^{k} \right )^{2} \right )^{d}$\\                                                                    &  \\
Linear splines kernel in one-dimension   & $k\left ( x,y \right ) = 1+xy+xy \, min\left ( x,y \right ) -\frac{x+y}{2} \,min \left ( x,y \right )^{2} +\frac{1}{3} \,min\left ( x,y \right )^{3}$\\   \\
\hline
\end{tabular}
\caption{Commonly used SVM Kernal Functions}
\label{KernalFormulas}
\end{table*}

\subsection{Convolutional Neural Networks (CNN)}

\begin{figure*}[h!]
\includegraphics[width=10cm]{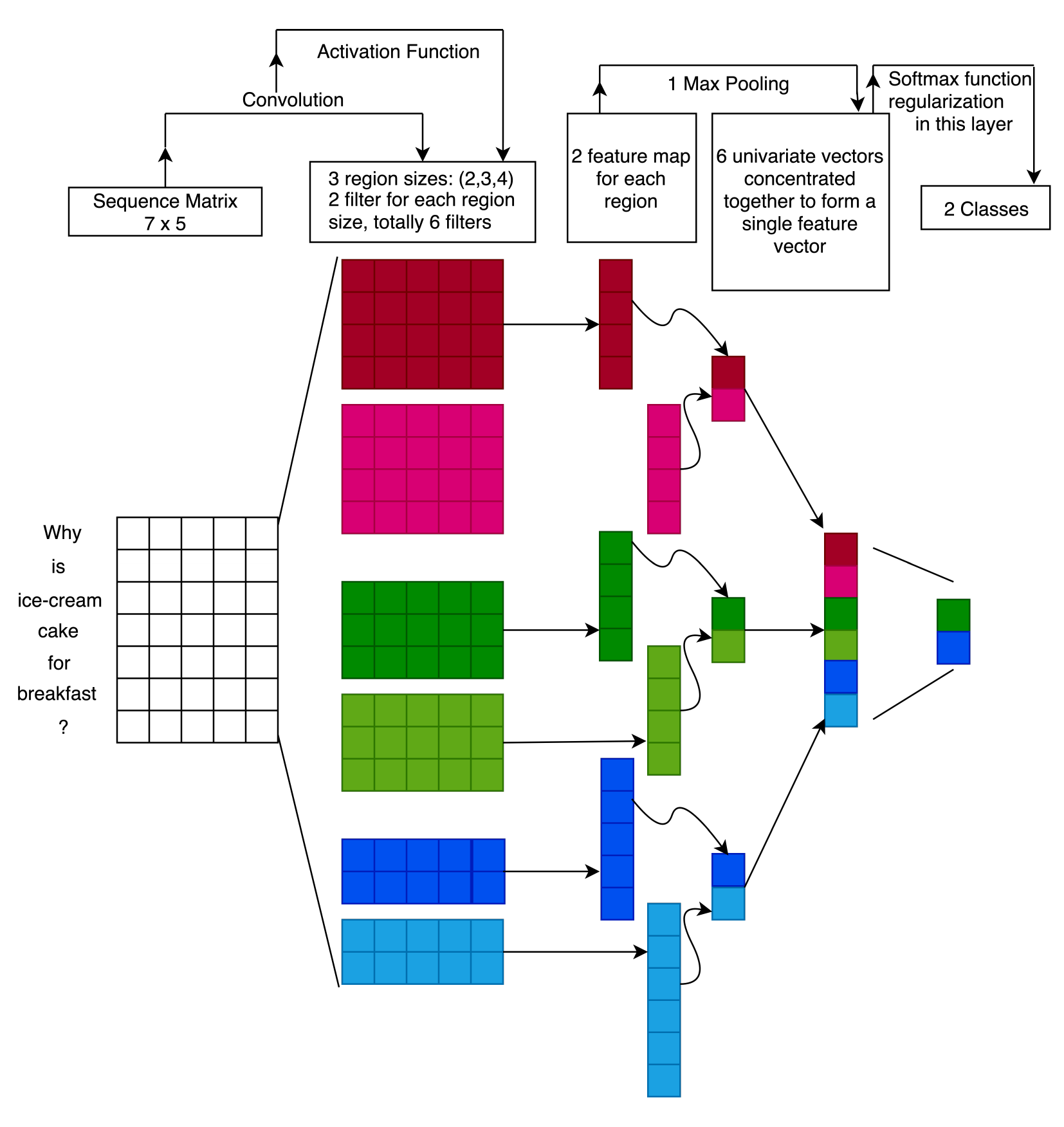}
\centering
\caption[CNN]{An Example Convolutional Neural Network (CNN) \cite{zhang2017}}
\label{CNN}
\end{figure*}

A CNN is a deep neural network architecture that rose to prominence for its applications in computer vision, but has been used for language modelling, spam detection, sentiment classification amongst other NLP tasks \cite{DosSantos2014} \cite{Kalchbrenner2014} \cite{Krizhevsky2017} \cite{Ren2017} \cite{Yin2017}. The distinctive feature of a CNN is the presence of convolution layers. The convolution layers serve to convolve the features by applying a sliding window or filter to the feature matrix using a non-linear activation function \cite{Ren2017} \cite{Chandrasekaran2022}. Unlike in computer vision, where the input data is usually pixels, in NLP tasks, sentences or full documents are represented as the input matrix \cite{Johnson2015}. Each row corresponds to a single token, usually a word vector derived from a word embedding technique or a one-hot encoded-word vocabulary index, although character embeddings may also be used \cite{DosSantos2014} \cite{Johnson2015} \cite{Ren2017} \cite{zhang2017}. Thus a 7-word sentence using 5-dimensional embedding would use a $7\times5$ input matrix. The sliding windows in NLP tasks are generally the same width as the embedding dimensions but the height can vary, thus, in our $7\times5$ input matrix example one would expect to find windows such as $3\times5$ or $2\times5$ as show in Fig. \ref{CNN} \cite{DosSantos2014} \cite{He_2015_CVPR} \cite{Kim2014} \cite{zhang2017}. 

Some relevant CNN hyper-parameters are padding, stride size, pooling layers and channels. The padding determines how the window is applied to edge elements of the input matrix. Wide convolution or zero padding fills the elements that fall outside the matrix with zero values. In contrast, narrow convolution does not use padding, and the window is bound by the edges of the matrix \cite{DosSantos2014}\cite{Kalchbrenner2014}. Stride size determines how many elements the window shifts in each step \cite{He_2015_CVPR}. Pooling layers, typically max-pooling in NLP tasks, are used to reduce the input matrix dimensions while producing a fixed output matrix, allowing for input matrices of various sizes \cite{Chen-2-2015} \cite{He_2015_CVPR} \cite{Johnson2015} \cite{Yin2017} \cite{Zeiler2014}. Channels reference different input types; for example, in NLP tasks, separate channels may be used for word2vec and GLOVe embeddings, or one channel may be kept static, while the other is tuned via back-propagation \cite{Liang2020} \cite{Vieira2017}. The output of each of the channel layers and filtered max-pooling layers are often concatenated to form a new feature vector that may then be passed into another CNN layer, a neural network or a softmax function \cite{Kalchbrenner2014} \cite{Ren2017} \cite{Vieira2017}.

\subsection{Recurrent Neural Networks (RNN)}

\begin{figure}[h]
\includegraphics[width=12cm]{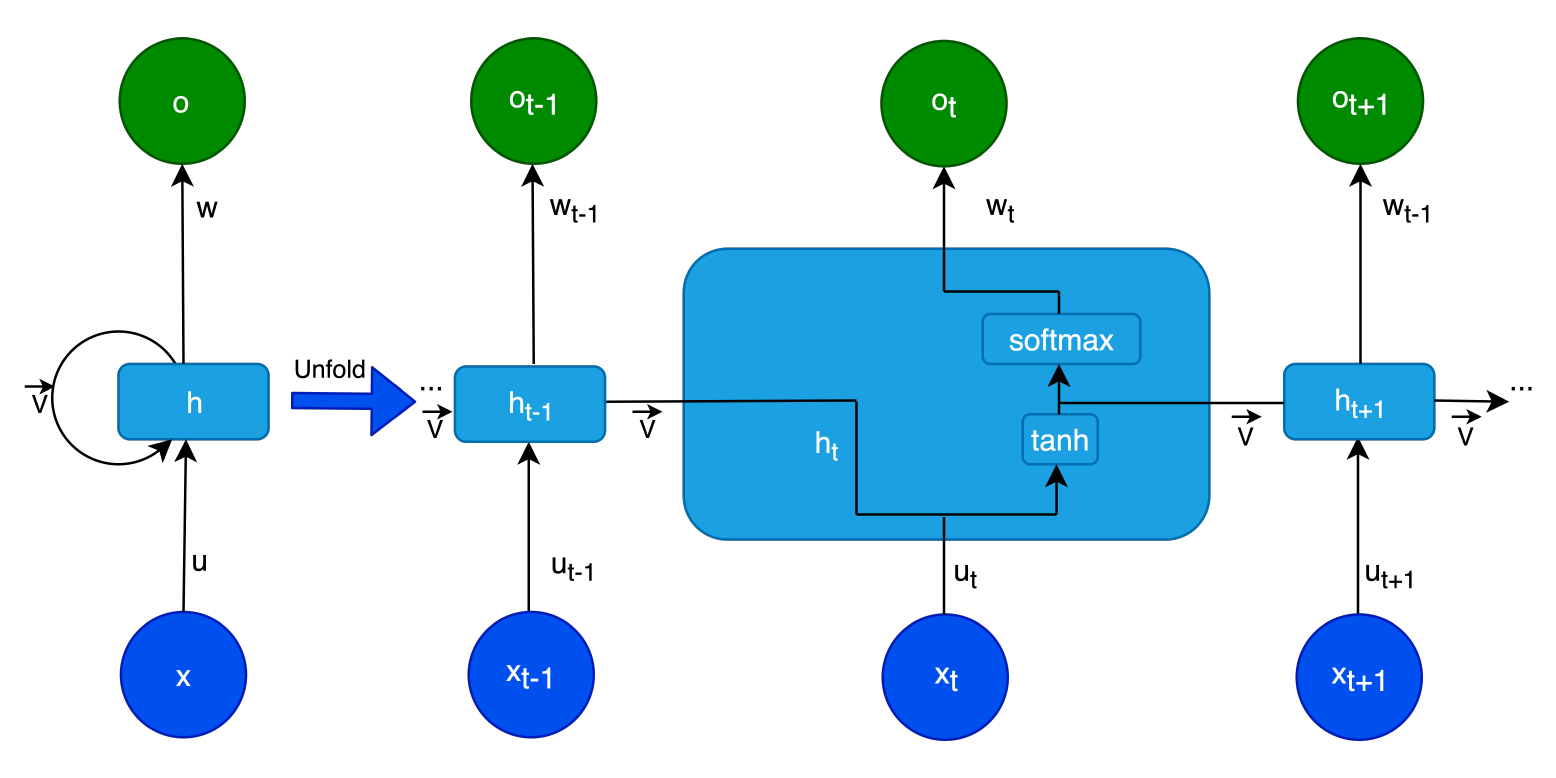}
\centering
\caption[RNN]{A Basic RNN Architecture and its Unfolded State}
\label{RNN-Unfold}
\end{figure}

RNNs are a deep learning architecture often used in NLP algorithms. Unlike traditional neural networks, where each element is considered independent from one another, RNN's sequentially processes each element and carries forward a memory or context of the prior elements. This makes it very useful in language modelling, text prediction, and machine translation \cite{cho_2014} \cite{Luong2015} \cite{Tani2019} \cite{Yin2017} \cite{Yu2019-2}. Fig. \ref{RNN-Unfold} displays a basic RNN architecture and its unfolded state. Unfolding an RNN allows for visualizing how each block or cell interacts with one another over time. $X$ would be the input for NLP tasks derived from a word embedding or one-hot encoded-word vocabulary index. $V$ is the memory vector that contains information from the previous elements. Generally, $V$ is initialized to an all-zero state so as not to provide memory to the first element. $H$ as shown in Fig \ref{RNN-Unfold} is the block or unit which combines the context memory vector $V$ with the current element input $X$, processed through a non-linear activation function, in this example $tanh$. The activation function output is passed to both the following elements block as the new $V$ memory vector and the output $O$, which is often processed with a softmax function.

Basic RNN architectures are susceptible to exploding and attenuating gradients \cite{Bengio1994}. Exploding gradients, which is when the gradients derived through back-propagation become very large, making the system unstable, are typically handled by gradient clipping, the practice of using a predefined gradient threshold \cite{Pascanu2012}. Attenuating gradients, also referred to as vanishing gradients, when the gradient itself diminishes to a near-zero value, making the system untrainable \cite{Bengio1994}. To compensate for attenuating gradients, two types of RNN models were developed and are commonly used for NLP purposes: Long Short Term Memory (LSTM) and Gated Recurrent Neural Network (GRNN) with much success \cite{Wang2016}.

\begin{figure}[ht]
\includegraphics[width=12cm]{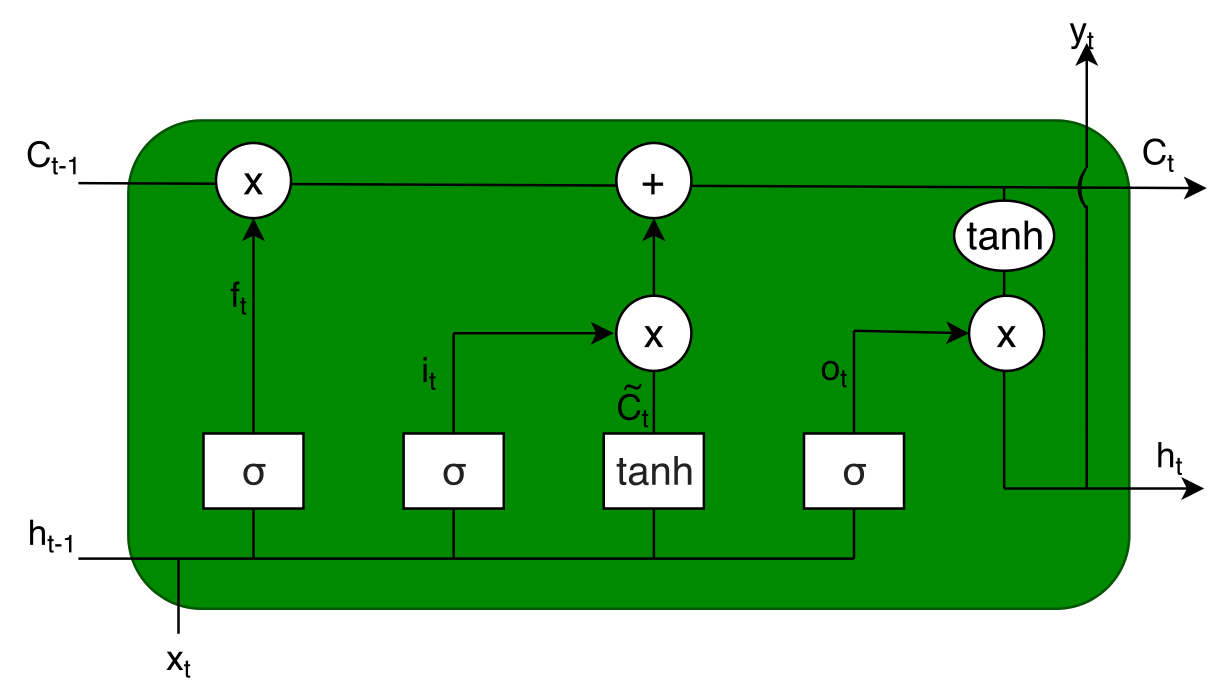}
\centering
\caption[LSTM]{Architecture of LSTM With a Forget Gate}
\label{LSTM}
\end{figure}

\begin{equation}\label{lstm1}
{\mathit{f_{t} = \sigma \left ( W_{f}\cdot \left [ h_{t-1}, x^{_{t}} \right ] + b_{f} \right )}}
\end{equation}

\begin{equation}\label{lstm2}
{\mathit{i_{t} = \sigma \left ( W_{i}\cdot \left [ h_{t-1}, x_{t} \right ] + b_{i} \right )}}
\end{equation}

\begin{equation}\label{lstm3}
{\mathit{\widetilde{C}_{t} =  tanh\left ( W_{C}\cdot \left [ h_{t-1}, x_{t} \right ] + b_{C} \right )}}
\end{equation}

\begin{equation}\label{lstm4}
{\mathit{o_{t} = \sigma \left ( W_{o}\cdot \left [ h_{t-1}, x_{t} \right ] + b_{o} \right )}}
\end{equation}

\begin{equation}\label{lstm5}
{\mathit{C_{t} =  f_{t} * C_{t-1}+i_{t}*\widetilde{C}_{t}}}
\end{equation}

\begin{equation}\label{lstm6}
{\mathit{h_{t} = o_{t}*tanh\left ( C_{t} \right )}}
\end{equation}

LSTM have recently been used to some success in sarcasm detection \cite{Ghosh2016} \cite{Ghosh2017} \cite{Huang2017}. LSTMs approach the challenge of exploding and attenuating gradients by establishing relationships between inputs to develop long-term dependencies via three control gates regulated by a $sigmoid$ function.The LSTM block architecture is more complex than the RNN block architecture. Referring to Fig. \ref{LSTM} there are three inputs to a LSTM block: $X\textsubscript{t}$, the input vector of the current element, for NLP purposes generally a word embedding or one--hot encoding; $C\textsubscript{t-1}$, the memory values from the previous block; and $h\textsubscript{t-1}$, the output from the previous block. $X\textsubscript{t}$ and $h\textsubscript{t-1}$ concatenate together to feed into the three sigmoid activation function gates: a forget, input and output gate; which will output values between zero and one \cite{YuYongSi2019}. 

The first gate, the forget gate, with output $f\textsubscript{t}$, equation (\ref{lstm1}), regulates the memory from the previous block, $C\textsubscript{t-1}$, as they undergo a pointwise multiplication operation. $f\textsubscript{t}$ values near zero will lead to forgotten memory from $C\textsubscript{t-1}$, while values near one will be retained memory \cite{YuYongSi2019}. The concatenated input values from $X\textsubscript{t}$ and $h\textsubscript{t-1}$ then pass through a $tanh$ operation  to produce candidate memory values $\widetilde{C}\textsubscript{t}$, equation(\ref{lstm3}), which is pointwise multiplied with $i\textsubscript{t}$, equation (\ref{lstm2}), the output from the input $sigmoid$ gate, to produce the new memory values. The new memory values are pointwise added to the result of $f\textsubscript{t} \cdot C\textsubscript{t-1}$. This creates the memory cell state  $C\textsubscript{t}$, equation (\ref{lstm5}), to be passed on to the next LSTM block.
This memory cell state $C\textsubscript{t}$ passes through a $tanh$ activation function which is pointwise multiplied with the output gates result $o\textsubscript{t}$, equation (\ref{lstm4}), which is used to regulate the output to both $h\textsubscript{t}$ (\ref{lstm6}), which is passed into the next LSTM block to be processed, and $Y\textsubscript{t}$, the output to the next layer or output function, usually softmax in NLP tasks.

\begin{figure}[h]
\includegraphics[width=12cm]{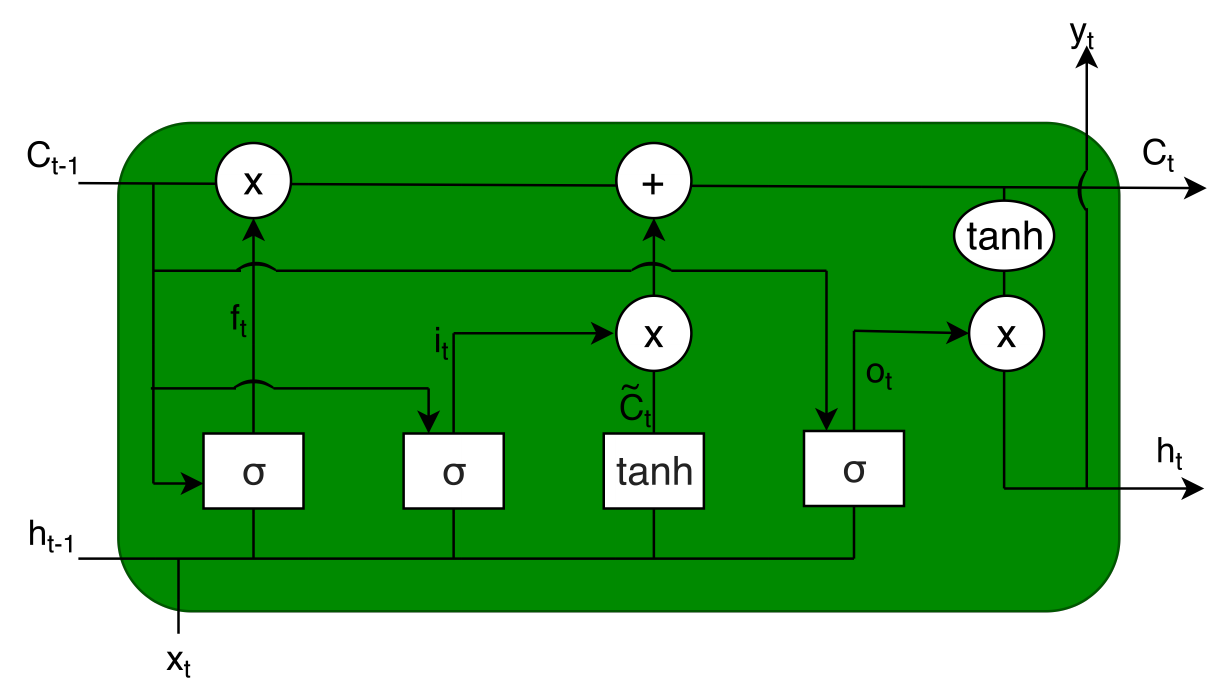}
\centering
\caption[peephole]{Architecture of LSTM With Peepholes}
\label{peephole}
\end{figure}

Some variants on LSTM architecture includes peepholes (Fig.\ref{peephole}) that are concatenated with the inputs to each gate before processing the sigmoid function \cite{YuYongSi2019}. Specialized LSTM variant networks have been developed such as attention-based LSTM, conditional LSTM, Bidirectional (BiLSTM) and Gated Recurrent Unit (GRU) \cite{chen_xu2017} \cite{HuangXY15} \cite{kadar2017} \cite{Wang2016} \cite{YuYongSi2019}.

\subsection{Word Embedding Techniques}
Word embedding is a technique for mapping real-value vector space representations of individual words \cite{Moya2017}. Words with similar meanings will have similar representations, which allows for grouping or clustering of like meaning words \cite{Kusner2015}. The word vector representations can be used as inputs into neural networks, which allows for use in deep learning algorithms \cite{Bojanowski2017} \cite{Lastra-Diaz2019}. Word embedding techniques generally fall into two categories; static and contextualized \cite{Giatsoglou2017}\cite{qudar2020}.

\begin{equation}\label{cos}
{ \mathit{\cos \Theta = \frac{A\cdot B}{\left \|A\right \|\cdot \left \| B \right \|}}}
\end{equation}
\subsubsection{Static Word Embedding}

\begin{figure}[h]
\includegraphics[width=12cm]{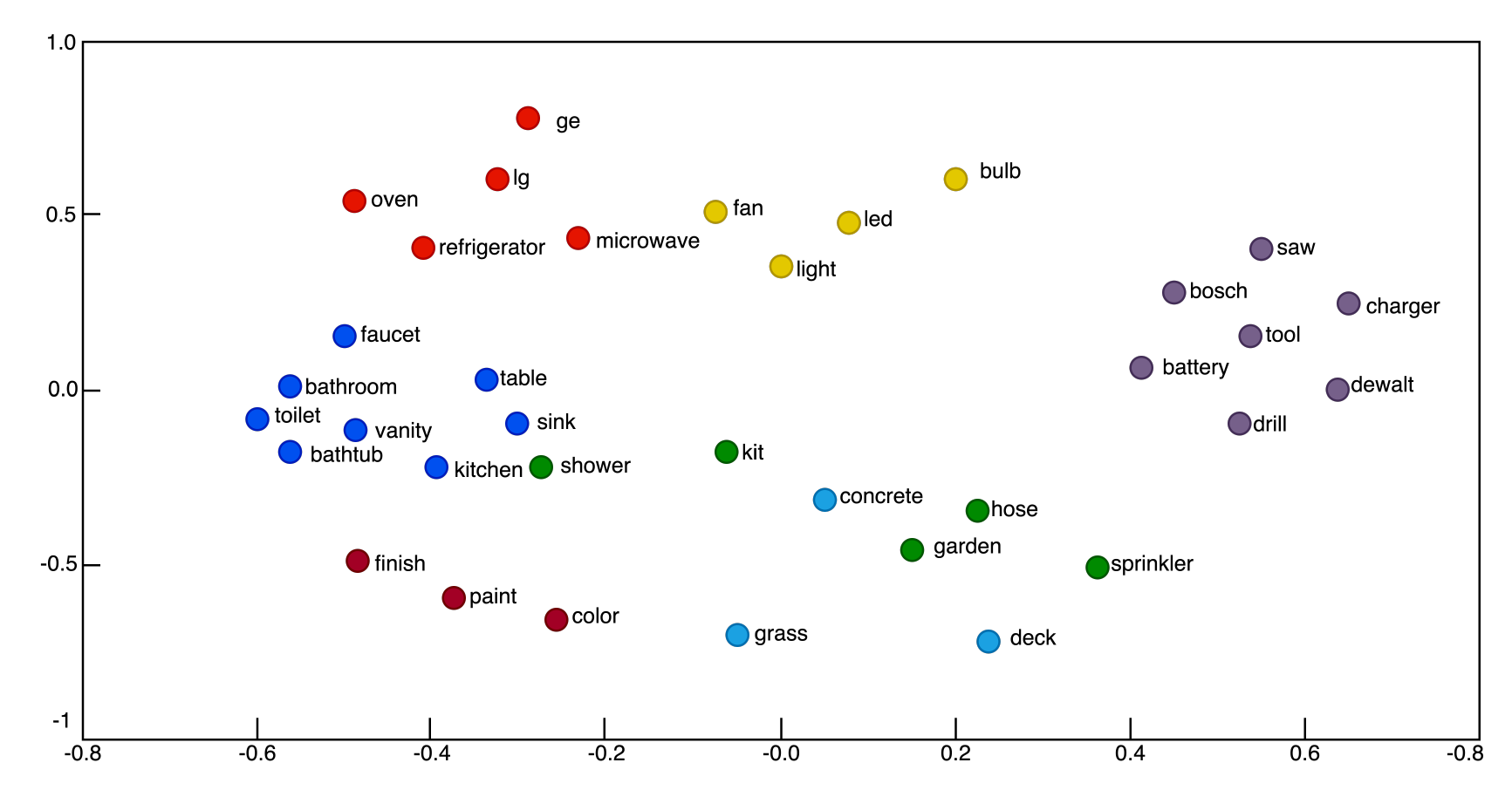}
\centering
\caption[2-D Word Embedding Representation]{2-D Word Embedding Representation} 

\label{2DWordEmbedding}
\end{figure}

Static word embedding, also known as semantic vector space or semantic modelling, is used in a variety of applications such as document clustering, sentiment analysis and text classification \cite{Finkelstein2015} \cite{Levy2014} \cite{Levy2015} \cite{Mikolov2013} \cite{Schnabel2015}. The mapped vector space representations for each word contain information about how each word is semantically used. These representations allow for grouping and classifying words \cite{Liu2015} \cite{Shang2018}. To illustrate; words like \textit{oven}, \textit{refrigerator} and \textit{microwave} would be grouped together for having similar word vector representations, while \textit{grass} would not, as its word vector model representation would be quite different (Fig. \ref{2DWordEmbedding}). Two frequently used methods are word2vec, and global vectors for word representation (GLOVe) \cite{Camacho-Collados2018}.

Word2vec maps a real-vector representation of each word by use in the words local context, meaning the semantic value is specifically dependent on the context derived from the surrounding words \cite{Chandrasekaran2021} \cite{Mitra2015} \cite{Pennington2014}. This mapping is done via a neural network with a single hidden layer and outputs a vector, which can then be fed into a deep learning algorithm if desired \cite{Chen2015} \cite{Mikolov2011}. Words with a high similarity, for example, using a cosine similarity function (Equation \ref{cos}), close to one, are similar, while those with low cosine similarities, close to zero, are dissimilar \cite{Mikolov2013}.
GLoVe addresses the local context limitation of word2vec by forming representations of the words using both local and global statistics \cite{Pennington2014}. GLoVe develops a local word vector, then further trains the representation using a constructed global co-occurrence matrix to derive a semantic relationship \cite{nguyen2015-2} \cite{Vaswani2017}.

\subsubsection{Contextualized Word Embedding}
\begin{figure*}[h!]
\includegraphics[width=10cm]{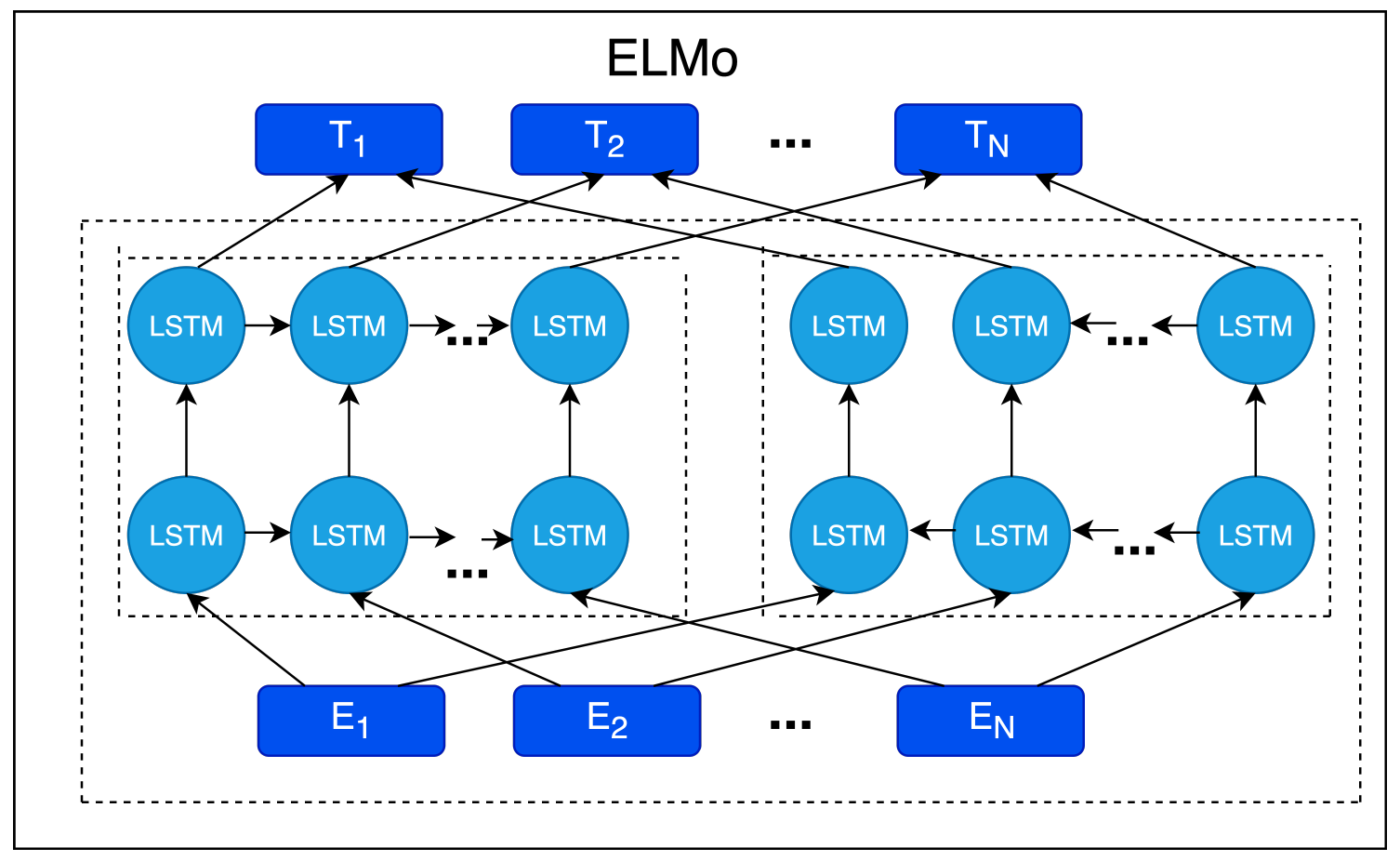}
\centering
\caption[ELMo Architecture]{ELMo Architecture }
\label{ELMo}
\end{figure*}

Contextualized word embedding techniques are used to estimate the probability distribution over a series of words to both provide context and distinguish from similar words \cite{Das2015}. Unlike static word embedding, which only produces vector representations, the final product from contextualized word embedding is vector representations and a trained model \cite{Wang2007}.  Additionally, conceptualized word embedding creates a separate vector representation for each different meaning of an individual word, determined by the context of how the word is used \cite{Das2015}. An example to illustrate would be with the sentences "I have an apple pie." and "I have an Apple watch." where Apple has a very different meaning in each sentence. Thus, each would have a different vector representation. Two prominent contextualized word embedding models are embeddings from language models (ELMo) and bidirectional encoder representations and transformers (BERT) methods. 

\begin{figure}[h]
\includegraphics[width=8cm]{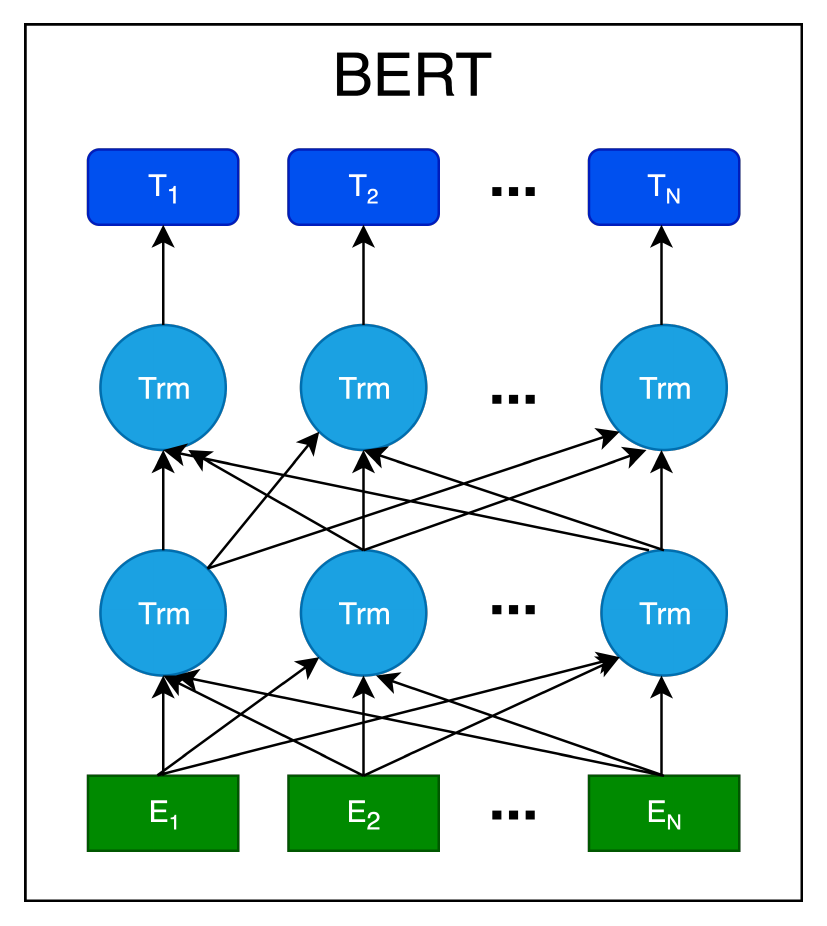}
\centering
\caption[BERT Architecture]{BERT Architecture}
\label{BERT}
\end{figure}

ELMo, whose architecture is shown in Fig. \ref{ELMo}, uses deep contextualized word representation features that take into effect the complex syntactic characteristics of the words such as context and word ambiguity when creating vector representations \cite{Kim2015} \cite{Mikolov2013-2} \cite{pittke2015} \cite{Wang2019}. These vector representations create separate representations for a word based on the context used \cite{pittke2015}. First, the representations are created by pre-training the bidirectional model on a large text corpus. The bi-directional model analyzes the words relevant to the surrounding words in both left to right and right to left manner and then concatenates the results to develop the context of how the word appears in the sentence \cite{Neumann2018}.
\begin{figure*}[h!]
\includegraphics[width=14cm]{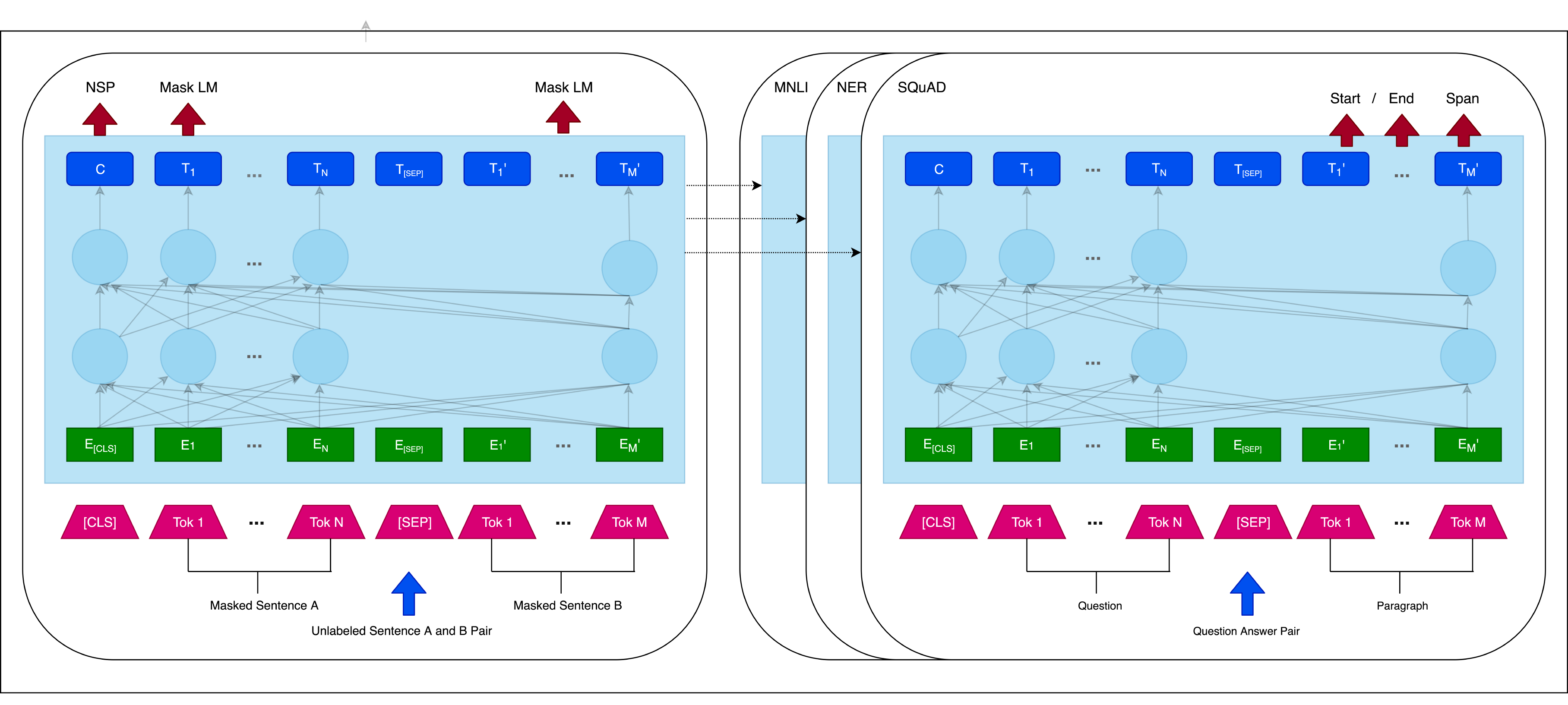}
\centering
\caption[BERT Pre-Training Architecture]{BERT Pre-Training Architecture }
\label{BERT_pre-train}
\end{figure*}

Unlike ELMo, which uses a feature-based approach to word representations,  BERT makes use of a pre-trained neural network \cite{Devlin2018}. Additionally, BERT uses a bi-directional transformer architecture (shown in Fig. \ref{BERT}), composed of vertically stacked transformers, which allows for parallel processing of words, unlike previous models that sequentially process words \cite{Vaswani2017}. BERT pre-trains using both masked sentence modelling (MLM) and next sentence prediction (NSP) techniques as shown in Fig. \ref{BERT_pre-train} \cite{Devlin2018}. MLM will choose random words and then conceal them. The model is then expected to predict the hidden words; this helps to avoid the issue in bidirectional models where the words can see themselves \cite{Devlin2018}. NSP provides the model with paired sentences to see if the model can predict if the second sentence comes after the first or not; this allows the model to learn how sentences interconnect \cite{Vaswani2017}.

\section{Summation of Modern Sarcasm Detection Models} \label{Summation of Modern Sarcasm Detection Models}

This review will perform a comparative analysis of high-quality articles derived from a Google Scholar search for "sarcasm detection" meeting the following criteria: 
\textit{
The article must come from a venue with at least a Scimago Journal \& Country Rank (SJR)\footnote{https://www.scimagojr.com/} of at least 50 and either Q1 or Q2, or Google Scholar\footnote{https://scholar.google.ca/} publication H5-Index of at least 50 or a minimum of at least 100 Google Scholar citations or a CORE\footnote{https://www.core.edu.au/index.php/} conference rating of at least A. Additionally, the articles will be recent and published no earlier than 2015}. 

The following exceptions will apply: \textit{Some articles are essential to defining or deconstructing the linguistic concept of sarcasm. They may come from earlier dates as this is not a novel concept, and the quality of the venue is more important than the date of publication. The results from the article "Sarcasm Detection on Czech and English Twitter \cite{Ptacek2014}", while published in 2014, are included as they are used by many of the articles as a comparative baseline. Some articles will be cited as a requirement for dataset referencing, as in the case of Sarcasm Corpus V2.}

\textit{Additionally, the Second Workshop on Figurative Language Processing hosted in July 202 by the Association for Computational Linguistics(ACL) included a sarcasm detection shared task \cite{ghosh-etal-2020-report}. While the workshop is too new of a venue to reach the consideration metrics noted previously, the ACL is a prestigious venue. As such, the top 3 scoring models \cite{lee2020} \cite{jaiswal2020} \cite{dong2020} from the Twitter dataset competition are included for analysis.
}

\subsection{Word Embedding Approaches}
Joshi \cite{Joshi2016-2} proposed an incremental improvement by allowing for word embedding to be used as additional features in previously assessed models of sarcasm detection. This improvement was proposed after noting that some forms of sarcasm may lack sentiment-laden words; thus, looking for potential sentiment inversion will not detect it. An example used in the paper is "A woman needs a man like a fish needs a bicycle." It is commonly known that fish do not use bicycles, which provides the context for the sentence. Using word vectors to calculate similarity scores, the similarity of man and woman is calculated as 0.766 while the similarity of fish and bicycle is 0.131. This semantic dissonance could be used to provide context to the sentence. Thus Joshi implemented an improvement to previous models by using the scores of the most similar and most dissimilar scores in a sentence as additional features.

The data set was created by using quotes from GoodReads\footnote{https://www.goodreads.com/} - a website for book recommendations that also allows for annotated quotations to be posted by users. 3,629 quotes were used for the dataset, with 759 user tagged as `sarcastic' and the rest tagged as `philosophy' on GoodReads. Any quote tagged as both was discarded. The unbalanced dataset was chosen as past experimentation reported on a similar skew \cite{Riloff2013}.

To develop the proposed model, Joshi combined the features from the works proposed by Liebrecht \cite{Liebrecht2014} (features consisted of unigrams, bigrams and trigrams), Gonzalez-Ibanez \cite{Gonzalez-Ibanez2011} (consisting of unigrams and dictionary derived features), Buschmeir \cite{Buschmeier2015} (features included unigrams and hyperbolic markers), and Joshi's \cite{Joshi2015} own earlier work (unigrams, implicit incongruity patterns and explicit incongruity patterns were used as features) and processed them using an SVM model and 5-fold cross-validation. The most similar and dissimilar word vector similarity scores were added to these features. The word vector similarity scores were calculated using four different word embeddings: LSA, GloVe, Dependency Weights and Word2Vec. The model was first analyzed without the word embedding features to provide a baseline for comparison. After implementing the features, it was determined that Word2Vec would provide the most significant average F1 score increase of 1.143\%.

\subsection{Sarcasm Detection through Context}
An analysis model developed by Ghosh \cite{Ghosh2018} sought to identify sarcasm by the conversational context using the Tweets previous to and following the target Tweet. This model differs from most methods that assess the post or Tweet in isolation to determine if it is sarcastic. Data was collected from three sources. The first source was The Sarcasm Corpus V2 subset of Internet Argument Corpus (IAC\textsubscript{v2}) containing sarcastic and non-sarcastic labelled posts. Secondly, the Self-Annotated Reddit Corpus (SARC) where sarcastic posts are labelled with a "/s" closing text, was used. Finally, a Twitter corpus was constructed by collecting author annotated Tweets via the Twitter developer APIs using self-labelling hashtags.

Sarcastic labelled Tweets were determined by using the hashtags \#sarcasm, \#sarcastic and \#irony and sincere Tweets by hashtags designed to label as non-sarcastic positive (such as \#happy, \#love, \#lucky) Tweets and non-sarcastic negative (such as \#sad, \#hate and \#angry) Tweets. The `reply to status' flag was used to determine if a Tweet was in response to another Tweet. If so, the previous Tweet was also downloaded to provide conversational context; if possible, entire threaded conversations were collected. A fourth corpus was derived from the IAC\textsubscript{v2} by determining if posts in the original IAC\textsubscript{v2} could be found as preceding posts within the dataset. This fourth corpus allowed for the creation of a data set where a post has both a preceding and succeeding post (IAC\textsubscript{v2}\textsuperscript{+}).

Features were derived from the posts using n-grams (unigram, bigram and trigrams), lexical features and sarcasm indicators. The lexical features were derived from the Linguistic Inquiry and Word Count (LIWC), Wordnet-Affect, and MPQA Sentiment Lexicon to categorize sentiment, negation, and 64 categories based on linguistic and physiological processes, personal concern and spoken categories. Sarcasm Markers included hyperbolic words, morpho-syntactic markers, and typography. Hyperbolic words were determined using the MPQA lexicon to identify strong subjective words common in sarcasm. Morpho-syntactic markers are identified by the use of more than one exclamation point, tag questions ("didn't you," "aren't we"), and interjections ("wow," "ouch") that occur commonly in ironic utterances. Typographic markers include capitalization, quotations, emotes and frequent punctuation marks.

Data was then divided into training, development and test datasets (80\% training/10\% development/10\% test). Models were then developed using combinations of TF-IDF, attention-based LSTM and conditional LSTM networks and either current post or current and previous post. Results on Twitter data showed that using LSTM with sentence-level attention analyzing both current and preceding Tweets earned the highest precision (77.25) and F1 (76.07) scores and second-highest recall (75.51), which was 1.02\% below the top precision result for sarcasm detection. When considering the ability to detect non-sarcastic Tweets, this model was within 3\% of the top-performing model for precision (72.65), recall (74.52) and F1 (73.57) scores.

\subsection{Sarcasm Detection by Classifier}
Ptacek \cite{Ptacek2014} proposed using a supervised machine learning approach to detect sarcasm in Tweets. This approach developed a model using SVM and MaxENt classifiers on a dataset of 780,000 English Tweets (130,000 sarcastic and 650,000 non-sarcastic) Tweets collected through the Twitter APIs. English Tweets were labelled as sarcastic if the hashtag \#sarcasm was included. Features were derived from the dataset via n-grams, lexical features (part of speech tags, high-frequency word patterns), and sarcasm indicators (emoticons, punctuation, word case). Both models were subjected to 5-fold cross-validation, and it was determined that the MaxEnt classifier performed better than the SVM for all feature set combinations. It should be noted that all of the F1 measures for both SVM and MaxEnt models fell into the low 90s for balanced and high 80s to low 90s for the imbalanced model. However, the confidence interval (CI) was exceedingly low, with all results having a CI falling between 14\% and 20\%.

Seeking to detect sarcasm through individual authors' historical Tweet behaviour, Rajadesingan proposed the SCUBA (Sarcasm Classification Using a Behavioural modelling Approach) model. This model would determine 355 features about the Tweet and author from the target Tweet and historical Tweets. The features were broken down into the following categories:

\begin{itemize}
 \item \textbf {Sarcasm as a Contrast of Sentiments: }This section included features based on sentiment scores and if there is any contrast in Tweet sentiment or author's historical sentiment.
 \item \textbf {Sarcasm as a Complex Form of Expression: }The features contained in this section include the number of words, number of syllables, syllables per word of the Tweet and variance in these features from the author's historical Tweets. 
 \item \textbf {Sarcasm as a Means of Conveying Emotion: }Features here were used to determine the mood, affect and frustration of the author by contrasting sentiment score trends in historical Tweets, frequency of Tweeting, Tweet word sentiment score distribution, historic sentiment score distribution, historic sentiment score range and time of day Tweet probabilities.
 \item \textbf {Sarcasm as a Possible Function of Familiarity: }
 This category contains features related to the author's grammar, diction, vocabulary, reading levels, and familiarity with sarcasm from historical Tweets. Additionally, features related to familiarity with the Twitter environment, such as usage familiarity (average daily Tweets, Tweet frequency, time difference between successive Tweets), Parlance familiarity (historical number of reTweets, mentions and hashtags, use of shortened words) and social familiarity (numbers of friends and followers and friends and followers divided by the authors' Twitter age) are located here. 
 \item \textbf {Sarcasm as a Form of Written Expression: }This section included typical sarcasm written markers, such as repeated letters, capitalization and punctuation as well as lexical density, number of intensifiers, pronoun use, and structural variation.
\end{itemize}

The data set was created via the use of the Twitter API and composed of 9,104 sarcastic annotated Tweets (determined by the presence of \#sarcasm or \#not hashtags) and an unstated number of non-sarcastic annotated Tweets. Up to 80 historic author Tweets were extracted, subject to availability, for feature construction. The SCUBA model then classified a Tweet as sarcastic or non-sarcastic using an SVM model subjected to 10-fold cross-validation, producing an accuracy of $0.9224$ in the best-tested scenario.

Eke \cite{eke2021} identified two flaws in earlier sarcasm models believed responsible for poor performance. Previous models ignored the text context or suffered from training data sparsity. The proposed Multi-feature Fusion Framework was designed to compensate for those flaws. The dataset was collected using the Twitter streaming API, farming English language Tweets between June and September of 2019. Tweets were annotated as sarcastic if the Tweet contained the hashtags \#sarcasm or \#sarcastic and non-sarcastic if the Tweet contained \#notsarcasm, \#notsarcastic or were devoid of hashtags. The resulting dataset of 29,931 Tweets contained 15,000 sarcastic and 14,931 non-sarcastic Tweets.
After preprocessing, the nine features types were extracted: lexical features, length of microblog, syntactic features, discourse features, emoticon features, hashtag features, semantic features and sentiment-related features. The classification was performed in two stages using five classifiers: decision tree, SVM, K-NN, logistic regression and random forest. The first stage made use of lexical features processed with Bag-of-Words (BoW) and term frequency inverse document frequency (TF-IDF) to form a dictionary. This dictionary is then run through the five classifiers to determine a prediction label based solely on lexical features. Random forest proved to be the best stage one classifier with a precision of $0.835$, recall of $0.832$ F1 of $0.832$ and accuracy of $0.832$. While random forest performed the best, it should be noted that all classifiers performed well. The lowest performing classifier was KNN with scores of  $0.784$, $0.776$, $0.774$, and $0.776$ representing precision, recall, F1 and accuracy accordingly.
The second stage used the dictionary created in stage one in conjunction with the remaining eight features groups and the prediction label. Using these fused features, the performance increased significantly with random forest maintaining its position as the top-performing classifier with a precision of $0.947$, recall of $0.946$, F1 of $0.946$ and accuracy of $0.945$  with the other four classifiers having performance scores falling within $0.03$ of the random forest results.

\subsection{Sarcasm Detection Through Deep Learning}
Ghosh \cite{Ghosh2016} approached automatic sarcasm detection by proposing a model involving CNN, LSTM and deep neural network (DNN) architectures. Two datasets were built by using the Twitter API. The first dataset used for training consisted of 18,000 sarcastic and 21,000 sincere Tweets. Sarcastic Tweets were determined by the presence of author annotation via hashtags including \#sarcasm and \#yeahright. The second dataset consisted of 2,000 Tweets annotated by an Internet team of researchers. The proposed model padded the initial input layer so that all inputs would be of equivalent size. The input word embedding dimension was determined to be 256 as attempts with higher dimensions did not produce better results but required more time to train. This input layer was fed into a 2-Layer CNN using a $sigmoid$ activation function before passing to a pooling layer. The pooling layer then passed into a 2-Layer LSTM network which fed directly into a fully connected DNN, which terminated in a softmax output. This model configuration was able to achieve an F1 of $0.921$. A recursive SVM approach was also used in the study but had significantly poorer results with a top F1 score of $0.732$. 

Seeking to improve on this model, Ghosh \cite{Ghosh2017} proposed a model that would include context and author state-of-mind combined with a methodology for reducing noise in the dataset. To provide context, only Tweets used in replies were analyzed, with the Tweet being responded to used for context. The authors' state-of-mind was determined by using the AnalyzeWords \cite{Tausczik2010} web service algorithm. AnalyzeWords processes a Twitter users last 1,000 words prior to the timestamp of a given Tweet and creates a broad strokes psychological portrait by providing a score of 0 to 100 in eleven dimensions: Upbeat, Worried, Angry, Depressed, Plugged in, Personable, Arrogant, Spacy, Analytic, Sensory, and In the Moment.

Data collection methodology was improved to reduce noise in labelling through the use of a Tweet author feedback system. Noting that high profile Twitter users are magnets for sarcastic Tweets, a Twitter bot was developed that would randomly choose Tweets addressed to any of the current 700 top Twitter users (determined at the time by the no longer hosted TwitterCounter.com). These Tweet authors would be contacted to ask if the Tweet was sarcastic or not. Once the author replied concerning the Tweet in question was sarcastic, the Tweet being replied to, the authors' response, and the 50 Tweets prior to the Tweet timestamp were all harvested and stored. This process allowed for building an author annotated dataset of Tweets without relying on hashtags for labelling. Using this methodology, a dataset of 40,000 Tweets was built, with 18,000 being non-sarcastic and 22,000 being sarcastic.

Building on the previously proposed model \cite{Ghosh2016}, Ghosh added a parallel input for the context Tweet. Both input Tweets would parse through a word embedding layer using Word2Vec then pass into independent 2-layer CNN's and a bi-directional LSTM (BLSTM) layer. The outputs of the BLSTM layers would concatenate together along with the output of the AnalyzeWords embedding layer. The AnalyzeWords embedding layer is fed via the 11 feature values derived from the AnalyzeWwords algorithm, performed on up to 1,000 words from the up to 50 Tweets made by the author prior to the timestamp on the current Tweet of interest, which is outputted in vector form. The concatenating layer then feeds into a DNN and finally outputs to a softmax layer. The model was able to produce an F-score of $0.90$.

A recent proposal by Son \cite{Son2019} made use of BLSTM, CNN and word embeddings in developing the sAtt-BLSTM sarcasm detection model. The sAtt-BLSTM model was trained and tested on two datasets. The first dataset was derived from the hashtag annotated SemeVal 1,015 task 11 dataset \cite{ghosh-etal-2015-semeval}, creating a relatively balanced dataset of 15,961 Tweets with 7,994 annotated as sarcastic and 7,324 annotated as non-sarcastic. The second dataset was imbalanced, containing 15,000 sarcastic and 25,000 non-sarcastic Tweets, derived from sampling random Tweets then parsing them through the Sarcasm Detector\footnote{http://thesarcasmdetector.com/} to annotate the Tweet as sarcastic or not.

The sAtt-BLSTM architecture feeds an input Tweet text into a GloVe word embedding layer to build a word vector table for the attention-based BLSTM layer. The BLSTM layer uses a softmax attention function to generate an output vector of high-value words that are then concatenated with an auxiliary feature vector derived from the Tweet. This feature vector includes potential sarcasm markers, including the number of exclamation points, question marks, periods, capitalized letters and use of the word "or." The concatenated vector then feeds into a CNN using a ReLU activation function and max-pooling output layer prior to flowing into a fully connected linear transformation and a softmax output layer which classifies the text as sarcastic or non-sarcastic.
The model was able to achieve an F-Score of $0.9160$ on the SemeVal dataset and $0.8828$ on the imbalanced random Tweet dataset.

Felbo \cite{Felbo2017} took a novel approach to sarcasm detection through emoji prediction, where the emojis are used to provide emotional context. While this model was developed to detect sentiment, emotion, and sarcasm, this research will focus on the sarcasm detection aspect. The training dataset used consisted of 1,246 million Tweets, from January 2013 through to June 2017, containing at least one of the sixty-four most common emojis. If a Tweet contains multiple common emojis, then the Tweet is replicated so that a copy of the Tweet would be created and labelled under each of the emojis it contains. The model, named DeepMoji by the authors, consisted of an embedding layer, two biLSTM layers, an Attention layer and outputs in a final Softmax layer. The embedding layer projected each word into a 256 dimensional vector space with a hyperbolic tan activation function to force a dimensional size constraint of between -1 and 1 on each dimension. This then passed into two biLSTM layers containing 512 hidden units per direction. The Attention layer then performed an algorithm on each word to weigh its importance, finally passing to the Softmax output layer. For sarcasm detection, the sarcasm dataset version of IAC and IAC\textsubscript{v2} were used, and the DeppMoji model produced an F1 score of $0.69$ and $0.75$ respectively.

Poria \cite{Poria2016} developed a model for sarcasm detection that used emotion, sentiment and personality features derived from a pre-trained convolutional neural network. Each feature category (emotion, sentiment and emotion) passes through its own CNN to derive relevant features from the text and then are integrated together in a fully connected layer before being output in a softmax layer. The feature CNNs were all trained on different data sources. The sentiment feature CNN was trained on Twitter data from Semival 2014 \cite{semeval2014}, the emotional feature CNN was trained on an emotional feature training corpus developed in 2007 by Aman \cite{Aman2007IdentifyingEO} and the personality CNN was trained on the corpus developed from essays by Matthews in 1999 \cite{matthews1999}.

The model was then assessed on three different datasets. Two twitter datasets were developed by Ptacek \cite{Ptacek2014} the first of which contains 50,000 sarcastic and 50,000 non-sarcastic Tweets, the second an unbalanced dataset of 25,000 sarcastic and 75,000 non-sarcastic Tweets. Tweets were labelled as sarcastic if they included \#Sarcasm as part of the text. The third dataset was Twitter data obtained from The Sarcasm Detector\footnote{http://thesarcasmdetector.com/} website, an unbalanced dataset of 20,000 sarcastic and 100,000 non-sarcastic Tweets. From The Sarcasm Detector dataset, a random sample of 10,000 sarcastic and 20,000 non-sarcastic Tweets was used to form the third dataset. When using a CNN-SVM classifier, the model was able to derive an F-1 score of $0.9771$ on the first dataset, $0.9480$ on the second dataset and $9330$ on the third dataset.

Seeking an alternative to models requiring pre-defined discrete features, Ren \cite{Ren2018} proposed two neural network models, CANN-KEY and CANN-ALL, both with an emphasis on contextual information from earlier Tweets. The CANN-KEY model seeks to eliminate redundant information in a Tweet and instead focus on keywords. The proposed CANN-KEY model makes use of a context-augmented neural network where two neural networks run in tandem. The first network is a CNN using the target Tweet for input and contains a single convolutional layer before passing into a pooling layer, using max, min, and average pooling techniques. The second neural network calculates the highest-scoring TF-IDF values from the contextual historical or conversational Tweets, which flows into a pooling layer using max, min, and average pooling functions. The pooling layers from both networks are then concatenated together in a hidden layer through a $tanh$ function before passing to an output layer.

The second proposed model, CANN-ALL, seeks to use all the information from the $X$ historical or conversational contextual Tweets in a six-layer neural network. This is performed by running $X+1$ CNNs in tandem then merging them for the final three layers. The first CNN uses the target Tweet as input, passing it through a CNN and then performing a max-pooling function in the pooling layers. The other $X$ CNNs have the same structure as the first, but each uses a different context Tweet for input. The vector representations of each Tweet formed in the pooling layer are then non-linearly combined in the hidden layer using a $tanh$ function before passing to a softmax layer and finally the output layer.

The dataset was reused by the author from a previous study on Twitter sentiment analysis using a word embedding \cite{ren_2016} containing 1,500 target Tweets, with an additional 6,774 Tweets providing historical context and 453 Tweets providing conversational context. No information on how the Tweets are annotated as sarcastic or sincere is provided. For evaluation 10-fold cross-validation was used and compared to a baseline derived from previous sarcasm detection models \cite{barbieri-saggion-2014-modelling} \cite{wang_ren_2015} performed on the same dataset. The highest macro averaged F-1 score of $0.6328$ resulted from the CAN-KEY using just historical context, beating out the top baseline model macro averaged F-1 score of $0.6032$ by over $0.03$. Theorizing that feature selection may be hampering conventional sarcasm detection models, and that allowing a neural network model to induce features automatically would prove to be more accurate, Zhang\cite{Zhang2016} proposed a neural model making use of a bi-directional gated recurrent neural network (GRNN). The proposed neural model is comprised of two input components: one for the target Tweet, the other for the historical context Tweets. The context component inputs up to 80 of the authors' previous Tweets, determines the individual words' tf-idf values and outputs the top K tf-idf value words. The input target Tweet is processed by the bi-directional GRNN and outputs a feature vector. This feature vector is concatenated with the output of the context component then processed via a tanh activation function before outputting as a softmax result.

The dataset used in testing the model was obtained from an earlier model proposed by Rajadesingan \cite{Rajadesingan2015} collected by using the Twitter API. 9,104 Tweets using the sarcastic marker hashtags \#sarcasm and \#not were labelled as sarcastic, and all others were labelled as non-sarcastic. The sarcastic marker hashtags were removed from the target Tweet and any historical context Tweets used. When limited to just the local Tweet, the model performed with an F-measure of $0.7936$. If contextual Tweets were also analyzed, the model performed with a much improved F-measure of $0.9074$. Taking a similar approach, Amir \cite{Amir2016} recognized that previous works with discrete features and context or author's emotional state provided through analysis of historic author works was time and resource inefficient. Thus, a model was proposed that would make use of induced features via deep learning and user embeddings. The model input sentence vector is initially processed through a word pre-trained embeddings matrix. This is passed through a CNN layer composed of 3 different sized filters, each using a ReLU activation function. The ReLU outputs are then pooled using max-pooling and concatenated together with the author's user embedding value. This concatenated feature map is then processed by a feature map ending in a softmax function which declares whether the function is sarcastic or not. The dataset used was provided by a fellow researcher in automatic sarcasm detection \cite{Bamman2015} with the modifications that additional Tweets were scraped for each user to provide historical context and that any user that did not have accessible historical Tweets was removed from the dataset. Using 10-fold cross-validation, the model achieved an accuracy of $0.872$.

Cai \cite{cai2019} proposed that Tweets containing images could not be accurately classified as sarcastic or not without including the image in the analysis and thus proposed a multi-modal hierarchical fusion model. The proposed model included three input modalities. The modalities used to derive feature vectors are text, image, and image attributes. The text modality used a Bi-LSTM architecture to derive raw text vectors by concatenating the hidden states in each step. These raw text vectors were then averaged into a guidance text vector. The image modality derived raw image vectors from a pre-trained ResNet model, which were averaged to create the guidance vector. The image attribute modality used a separate pre-trained ResNet model to predict five attributes from the image, which were then through a GloVe embedding to create five raw attribute vectors, which were averaged to create the attribute guidance vector. The feature vectors from each modality were derived from the associated raw and guidance vectors, then passed through a two-layer fully connected neural network to classify if the Tweet was sarcastic or not. The dataset was composed of mined Tweets with images; if special hashtags were included (\#sarcasm is the only example provided by the author), then the Tweet was annotated as sarcastic. If the special hashtags were not present, the Tweet was annotated as non-sarcastic. The resulting model produced an F1 of $0.8018$, precision of $0.7657$, recall of $0.8415$ and accuracy of $0.8344$. However, the model is limited to Tweets that include images. Additionally,  the dataset may be biased as the author does not provide a complete list of the special hashtags, and they do not appear to be removed during cleaning. 

Son et al. proposed a deep learning model capable of handling sarcasm detection in real-time \cite{ Son2019}. The model named sAtt-BLSTM convNet, short for soft attention-based bidirectional long short-term memory and convolution neural network, is comprised of the following layers. The initial input layer feeds a Tweet into a GloVe layer, which maps each word in the Tweet into a low dimension embedding vector. These vectors then pass through a Bi-LSTM layer to determine the high-level features. These selected high-level features pass through a softmax attention layer to form the feature vector. The feature vector is concatenated with an auxiliary feature vector. The auxiliary feature vector consists of the number of incidents of the following pragmatic markers: exclamation marks, question marks, periods, capital letters and the word "or". This concatenated vector then feeds into a convolutional layer followed by a ReLU layer before passing through a fully connected softmax layer which categorizes the Tweet as sarcastic or not. When trained and tested on the SemEval 2015 task 11 dataset \cite{ghosh-etal-2015-semeval} the model performed admirably with an accuracy of $0.9787$, recall of $0.9683$, precision of $0.9214$ and F1 of $0.9357$. The high performance reached by this model may be due to the combination of induced features with pragmatic markers.

\subsection{Sarcasm Detection Using Transformers}
Team miroblog \cite{lee2020} proposed a model that won the Second Workshop on Figurative Language Processing's Sarcasm Detection Shared Task. The model named the Contextual Response Augmentation(CRA) was comprised of a 24-layer BERT architecture stacked with BiLSTM and NextVLAD \cite{ Lin2018} pooling layers. The dataset provided consisted of 5,000 training Tweets, 1,000 of which were reserved for validation. To provide a more extensive training set, team miroblog simulated additional training data. The Tweets were author annotated using hashtags. Sarcastic Tweets contained the hashtag \#sarcasm or \#sarcastic. Non-sarcastic Tweets were determined by matching hashtags with the Tweet sentiment. For example, Tweets with the hashtags \#happy, \#love, or \#lucky would have a positive sentiment. Conversely, Tweets containing the hashtags \#sad, \#hate, and \#angry would have a negative sentiment. 

Dialogue threads preceding the Tweets were used for creating context. Additional data points were created using the provided labelled dataset and freshly mined unlabeled dialogue threads. The labelled data was used to develop additional training samples via back-translation, the process of translating text to another language and then back to the original language. The languages used in back-translation were French, Spanish and Dutch. The unlabeled dialogue threads were used to generate samples via next sentence prediction using BERT. The architecture produced a precision of $0.932$, recall of $0.936$, and F1 of $0.931$, a significant increase over the second-place model.

Placing second in the 2020 shared task on sarcasm detection, team nclabj \cite{jaiswal2020} proposed a model making use of a RoBERTa \cite{liu2019roberta} infrastructure. RoBERTa is a modified pre-training approach to BERT modelling that allows for larger batch sizes, more epochs and far larger datasets. RoBERTa allows for byte-level BPE vocabulary and dynamic masking instead of character-level vocabulary and static masking as used in typical BERT models \cite{jaiswal2020}. The RoBERTa architecture made use of a pre-trained library then fed into three densely packed, fully connected layers before being categorized by a 2-way softmax layer. 
The training was performed using the provided dataset of 5,000 Tweets augmented with the most recent historical Tweets for each Tweet, if available, to provide context. The resulting model produced n precision of $0.790$, recall of $0.792$, and F1 of $0.790$

The final model of note from the Second Workshop on Figurative Language Processing's sarcasm detection shared task, submitted by team andy3223 \cite{dong2020}, also made use of a transformer architecture. Team andy3223 approached the task in 2 ways: using the individual Tweets in isolation and using them with mined historical dialogue threads for each Tweet to provide context. The model presented involved a transformer encoder feeding into a linear encoder. For the transformer encoder, three types of transformers were used:  BERT, RoBERTa, and ALBERT \cite{Lan2019}, a lightweight BERT architecture that uses parameter reduction techniques to improve model size scaling. 
Team andy3223 initially assessed the Tweets in isolation and determined the RoBERTa model provided the  F1 score. The RoBERTa model was then further developed using historical Tweets to provide context. The fully developed model produced a precision of $0.784$, recall of $.789$ and F1 of $0.783$.

Potamias \cite{Potamias2020} proposed a transformer-based model for sarcasm detection. The model's architecture consisted of a Recurrent CNN feeding into a pre-trained RoBERTa model, producing the biLSTM layer's input. The Bi-LSTM outputs are concatenated prior to entering a pooling layer, and finally, a softmax layer determines if the text is sarcastic. The RCNN and pre-trained RoBERTa model is used to capture contextual information. Supporting the RoBERTa model, the RNN layer is used to determine dependencies within the RoBERTa word embeddings. The max-pooling layer is used to convolve the LSTM output into a 1D-layer, with the final softmax layer determining if the Tweet is sarcastic or not. Using the Tweet datasets provided from the 2018 Semantic Evaluation Workshop Task 3 \cite{van2018}, this model was able to achieve an accuracy of 0.79, precision of 0.78, and F1 score of 0.78. 

Babanejad et al. proposed two sarcasm detection models using a modified BERT architecture to incorporate both contextual and affective features \cite{babanejad2020}. The resulting models were named ACE 1 and ACE 2, with ACE being an acronym for Affective and Contextual Embeddings. Both models performed well, with ACE 1 having slightly better results achieving an F1 score of $0.8457$ on the SemEval 2018 Twitter dataset \cite{van2018} and $0.9314$ on the IAC dataset \cite{IAC}. As ACE 1 provided superior results, it shall be the focus of this survey. The ACE 1 model is separated into the affective feature embedding (AFE) and contextual feature embedding (CFE) components.
The AFE was pre-trained using an unlabelled training corpus followed by an annotated sarcasm dataset for fine-tuning. This component comprises an affective feature vector representation section, a Bi-LSTM section, and the multi-head attention section. The affective feature vector representation section is where the input sentence has the emotion words in the sentence extracted. Using the NRC Emotion Intensity Lexicon \cite{Mohammad2017} each emotion word is given an intensity score, from 0 to 1, for the four underlying emotions of anger, fear, sadness, and joy. Two more scores are added using the NRC Emotion Lexicon \cite{ mohammad2012} to represent positive and negative sentiment. The affective feature vector for the sentence is calculated by averaging the affective feature vectors for each word multiplied by the word frequency for each emotion or sentiment. Additionally, this section is responsible for measuring the average emotion similarity score between words in the input text to create the emotion similarity feature representation vectors. The effective vector representation for a sentence is created by averaging the vectors for all words in the sentence. The effective vector representations are then passed through the BiLSTM section to encode the affect changing information of the sentence sequence, forming a matrix of hidden state vectors to pass to the input of the multi-head attention section. The Multi-head Attention Layer is used to capture the importance of the hidden affective features learned in the Bi-LSTM section, forming a global sequence representation matrix as the final output from the AFE component. The CFE component starts with a Large-uncased BERT architecture, trained on an unlabeled text corpus for MLM and NSP to create the embedding vectors. Next, a second BERT model is modified to include the affective features by training the BERT model using the AFE component output as the model's input for both MLM and NSP. The contextual embeddings for both BERT models are then concatenated together and combined with the AFE outputs with a fully connected layer leading into a softmax categorizer.

\subsection{Sarcasm Detection Through Lexical Analysis}
Bamman \cite{Bamman2015} noted that sarcasm is often accompanied by extra-linguistic cues in addition to linguistic or lexical indicators. Thus a model was proposed that included extra-linguistic cues in the analysis. Extra-linguistic cues used as features included intensifiers ("so," "too," "very") and abnormal capitalization (words with all letters capitalized or multiple words having the first letter capitalized). The dataset was composed of a sample of 19,534 Tweets from August 2013 to July 14, 2014. The sample was evenly split into 9,767 sarcastic and 9,767 non-sarcastic Tweets. Sarcastic Tweets were labelled as such if they met all of the following restrictions: the Tweet was at least three terms in length, written in English, not a retweet, not a response to another Tweet, and author annotated with \#sarcasm or \#sarcastic as the final term. Logistic regression was applied with the best result yielding an accuracy of 84.3\% and a CI of 95\%, but this included an analysis of audience response and authors' historic Tweets. When analyzing Tweets in isolation, the accuracy was 75.4\% with a CI of 95\%.

Bouazizi \cite{Bouazizi2016} took a different lexical approach in sarcasm detection by recognizing many sarcastic phrases contained similar lexical patterns and thus included pattern-related features to the sarcasm detection model. The datasets used were obtained through the Twitter API with sarcastic Tweets containing the hashtag \#sarcasm. The training set consisted of 6,000 Tweets, half of which were annotated as sarcastic. This set was also manually annotated and ranked on a scale of 1 to 6 (from highly non-sarcastic to highly sarcastic). The optimization set contained 1,128 sarcastic and non-sarcastic Tweets; these were not manually cleaned and purposely left noisy. The optimization dataset was still cleaned by removing non-English Tweets and Tweets with less than three words. An assumption was made that the sarcastic context may be found as a visual cue in the link or image that this model is not designed to assess; thus, Tweets containing images or links were also removed. The test dataset contained 500 sarcastic and non-sarcastic Tweets, which were manually classified as sarcastic or not sarcastic. None of the sets contained Tweets from the other sets. The features were processed using SVM, Random forest, MaxEnt and k-NN classification techniques. The best result earned an F1 score of 81.3\% using the random forest classifier. The detection model analysed four types of features: 
\begin{itemize}
 \item \textbf {Sentiment-Related} derived numeric features by rated words based on emotional strength (-1 to -5 for barely negative to very negative and 1 - 5 for barely positive to very positive) and then tallied the negative and positive scores for the Tweet. A count of positive, negative and possibly sarcastic emoticons was also derived from the Tweet and used as a sentiment feature.
 \item \textbf {Punctuation-Related} includes features related to the presence and number of various punctuation marks, including exclamation marks, question marks, dots, quotations, and full capitalization.
 \item \textbf {Syntactic and Semantic} includes features based on the number of uncommon words, the existence of common sarcastic expressions, the number of interjections and the presence of laughing expressions, such as "LOL" for laughing out loud.
 \item \textbf {Pattern-Related} features include commonly found lexical patterns in sarcastic statements like [NOUN PRONOUN \textit{be crazy}]. These patterns had to occur at least twice in the training data to be considered valid for use.
\end{itemize}
The features were processed using SVM, Random forest, MaxEnt and k-NN classification techniques. The best result earned an F1 score of 81.3\% using the random forest classifier. 

Joshi \cite{Joshi2015} aimed for a more linguistic feature-focused approach than earlier sarcasm attempts had used. Building off the research performed by Riloff \cite{Riloff2013}, Joshi wanted to include features derived from context incongruity directly instead of tangentially with the context incongruity broken down into explicit and implicit incongruity. Explicit incongruity is when words of opposing sentiment are used to invert the surface meaning of the text. Implicit incongruity is when more subtle and often requires a shared context, as in the phrase `I love Mondays' and generally consists of a positive verb combined with a negative noun. Four feature categories were implemented: lexical, pragmatic, explicit incongruity, and implicit incongruity. Lexical features were derived using feature selection techniques on unigrams, and pragmatic features included emoticons, expressions of laughter, punctuation and capitalization. Explicit incongruity features included the number of sentiment incongruities, largest positive or negative sequence, number of positive and negative words and Tweet lexical polarity. Implicit incongruity features were derived by modifying the rule base algorithm proposed by Riloff \cite{Riloff2013}. Three datasets were used for evaluating the model: Tweet-A was an unbalanced dataset of 5208 Tweets, 4170 of which were sarcastic. Tweet-A was derived through twitter crawling and labelled through author annotation. Tweets containing the hashtag \#sarcastic and \#sarcasm were labelled as sarcastic, while Tweets containing \#notsarcasm and \#notsarcastic were labelled non-sarcastic. Internal researchers manually overlooked the dataset to look for false labelled Tweets as additional quality control. Tweet-B, used by Riloff \cite{Riloff2013} in his earlier work, was the dataset of 2,278 Tweets, 506 of which were labelled sarcastic. Discussion-A was a dataset of 1,502 forum posts, with 752 labelled as sarcastic, randomly selected from the IAC\footnote{https://nlds.soe.ucsc.edu/iac}. Using all features, the model achieved an F1 score of 0.8876 on the Tweet-A dataset.

Taking a novel approach, Ren \cite{REN2020} sought to create a model to find a contrast between sentiment and situation for use in sarcasm detection. The proposed model is composed of LSTM encoders, two intra-attention memory networks and a CNN. The study used IAC-v1 and IAC-v2, as well as the Twitter dataset provided by Ptacek in an earlier sarcasm detection study \cite{Ptacek2014}. The first LSTM encoder is used for sentiment selection, with its output feeding into the first-level memory network. This memory network is used to capture the sentiment semantics. The second LSTM encoder encodes the full text as the second-level memory network and the CNN input. The second-level memory network is used to determine the contrast between sentiments or sentiment and situation. The CNN is used to create local information features for the given text; for this purpose, a local max-pooling layer is used as a traditional max-pooling layer could result in the loss of essential data. The results of the two memory networks and the CNN are then concatenated together to form the final feature representation, which is passed through a softmax sarcasm detection layer. On assessing the Tweet dataset, the model resulted in a recall of $0.8924$ and F1 of $0.8713$.

\section{Concluding Remarks} \label{Concluding Remarks}
Objective comparison of the different techniques is not feasible for various factors. First, there is no consensus on evaluation metrics. While some studies provide a variety of metrics, others have only published the most favourable. Secondly, datasets used by each study are different, coming from a large variety of data sources with varying annotation criteria. Even when some studies used the same Twitter reference datasets, many of the Tweets referenced may no longer be available in subsequent studies upon Tweet hydration. Third, some of the methodologies for dataset annotation are questionable. An example of this is using the sarcasm detector\footnote{http://thesarcasmdetector.com/} to determine if the post is sarcastic or not, as any error in the output of this tool will affect the integrity of the dataset.
\begin{figure}[h]
\includegraphics[width=14cm]{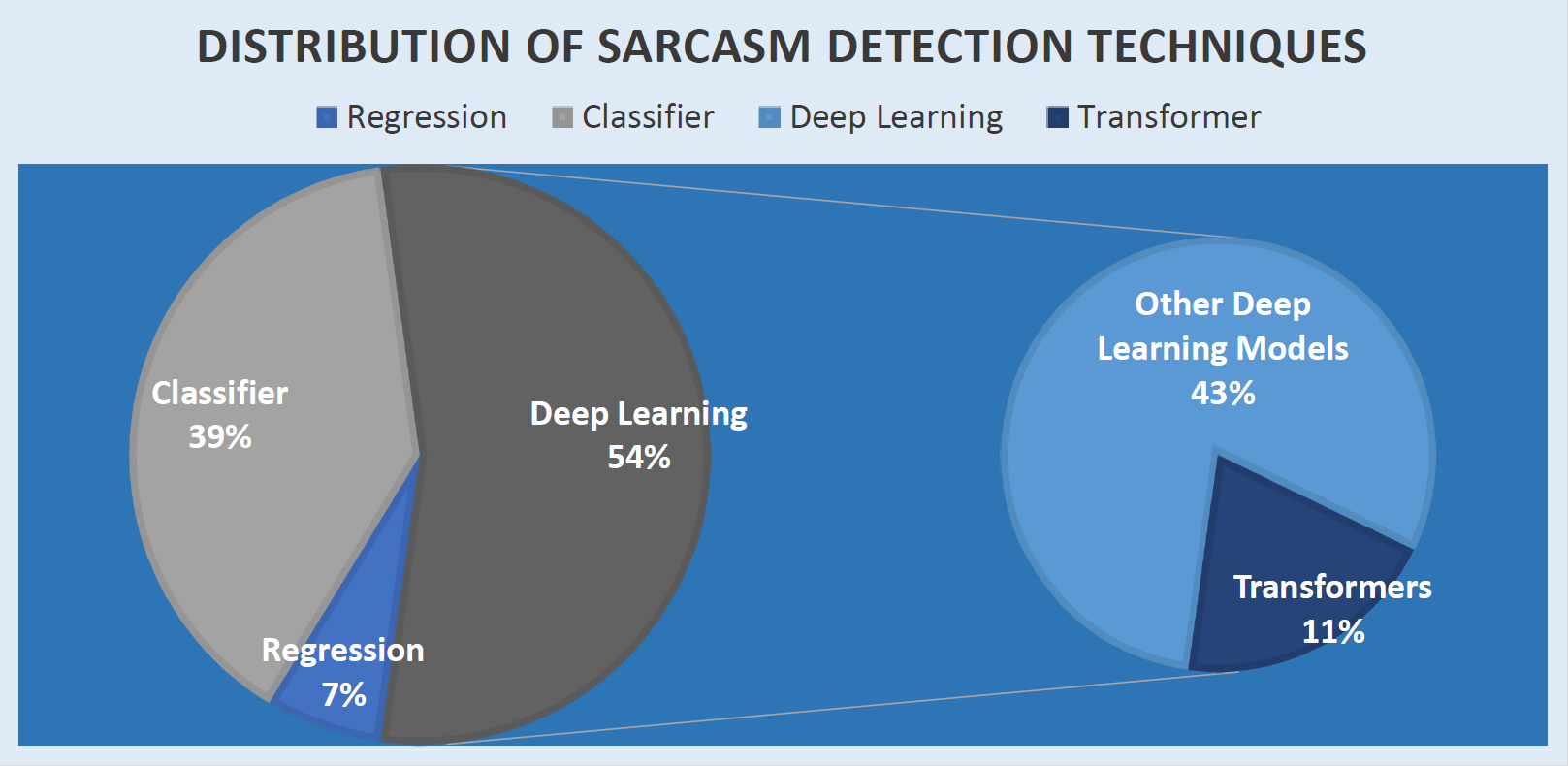}
\centering
\caption{A High Level Overview of Surveyed Sarcasm Detection Modelling Techniques}
\label{highlevel}
\end{figure}

\begin{figure}[h]
\includegraphics[width=14cm]{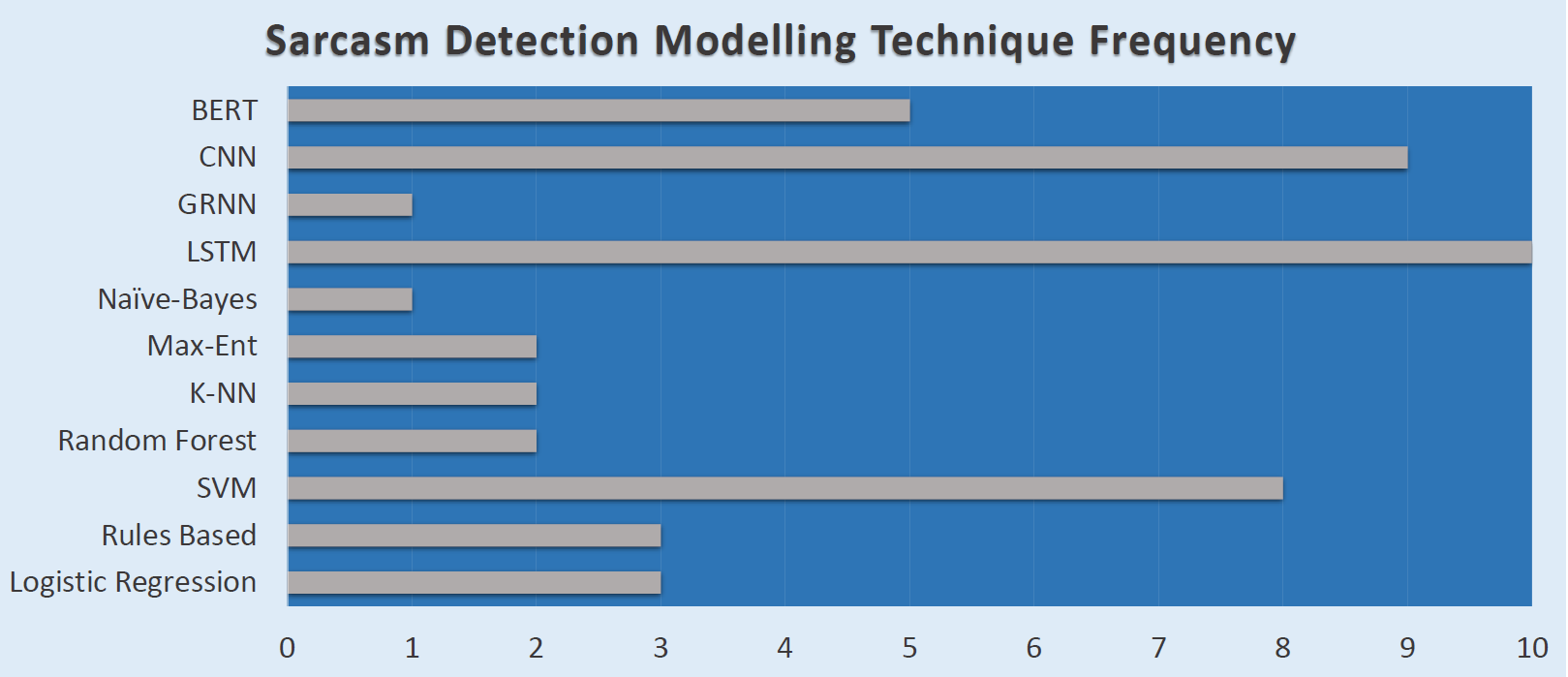}
\centering
\caption{Frequency of Modelling Usage in Surveyed Sarcasm Detection Models}
\label{techniquefrequency}
\end{figure}

\begin{figure}[h]
\includegraphics[width=14cm]{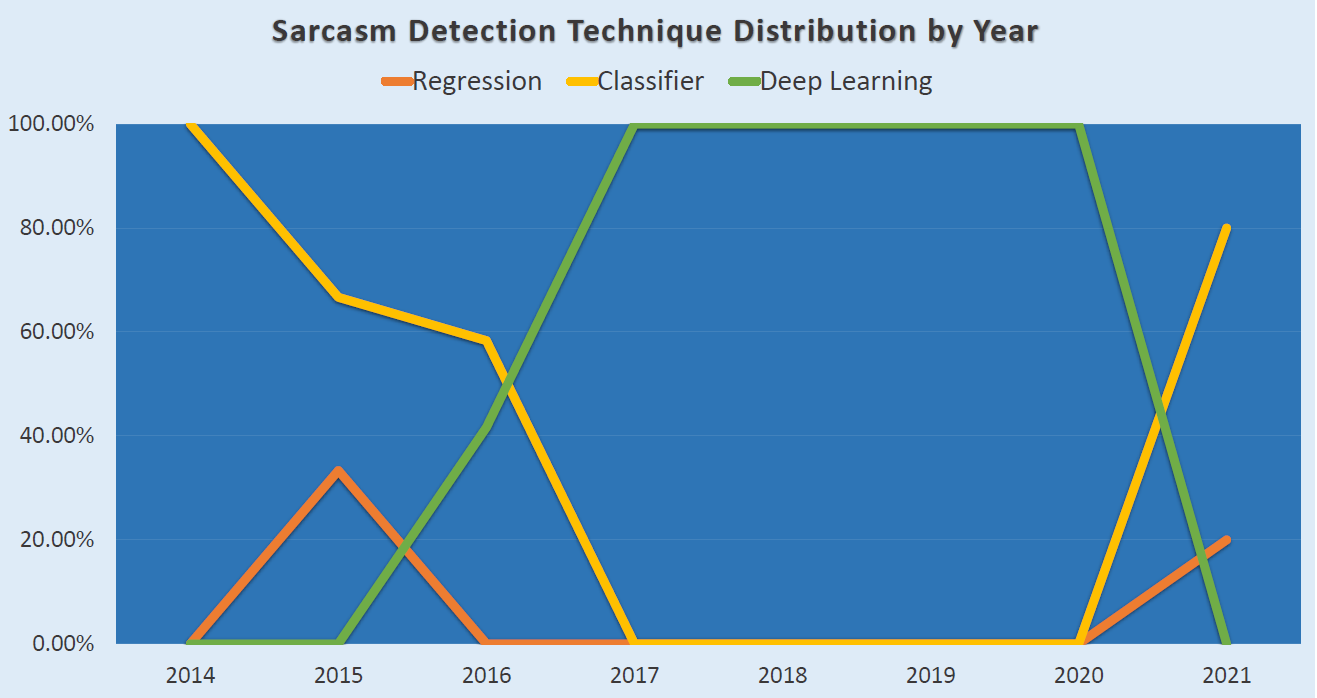}
\centering
\caption{Modelling Technique Distribution Over Time in Surveyed Sarcasm Detection Models}
\label{techniquedistribution}
\end{figure}

That said, a comparison of techniques used and how they have changed over time is of interest. In the studies analyzed, logistic regression modelling is in the minority, while over half of the models are deep learning, as shown in Figure \ref{highlevel}. Looking further, Figure \ref{techniquefrequency} displays that the dominant machine learning classifier is SVM. In contrast, the deep learning architectures used are predominately CNN and LSTM models, with transformer architecture in third. It should be noted that transformers are a relatively recent innovation that has been highly represented in recent sarcasm detection models. A further point of interest is highlighted in Figure \ref{techniquedistribution} where the rapid rise in Deep Learning technique usage coincides with a dramatic drop in classifier technique usage between 2015 and 2017. The exception is in 2021, where classifiers and regression techniques come back. It should be noted that this comeback is due to the use of four different classifier models and a regression model in Eke et al.'s architecture \cite{eke2021}. This architecture, combined with the research for this paper concluding in 2021, reducing the availability of high quality publications on the topic (as detailed in Section \ref{Summation of Modern Sarcasm Detection Models}), has artificially inflated the use of classifiers and regression compared to deep learning models in 2021.

An essential factor highlighted by these studies is that there are still some significant obstacles in automated sarcasm detection. Lack of shared background knowledge, long-form posts, use of numerals, and poor annotation all provide unique challenges. Shared background knowledge to provide context, which may be missing from the text, is often a pivotal component to sarcasm identification \cite{Ghosh2018} \cite{Joshi2017} \cite{Joshi2015} \cite{Schifanella2016}. This causes issues as models are unaware of this shared knowledge leading to classification error. "I thought that exam was a piece of cake" would be difficult to classify as sarcastic without knowledge of the difficulty of the exam, for example.

While generally not a particular issue for Twitter, long-form posts provide additional challenges in sarcasm detection \cite{Ghosh2018} \cite{Joshi2015} \cite{Zhang2016}. This was noted when long-form posts found in the Sarcasm Corpus V2, SARC and IAC the sarcasm tag denoted that the post was sarcastic but not which sentences were sarcastic. This becomes a problem when the post length is more than five sentences, as LSTM models have difficulty discerning if the post is sarcastic \cite{Ghosh2018}.

Context involving numerals is an added challenge. This is common on social media and extremely hard to detect without background knowledge \cite{Ghosh2018} \cite{Joshi2015}. An example of this would be "Happy birthday to my mother who just turned 29 this year!" from a poster that is obviously too old for this to be true.

Possibly the largest ongoing issue with automatic sarcasm detection comes from stems from poor dataset annotation \cite{Chaudhari2018} \cite{Felbo2017} \cite{Joshi2017} \cite{Wallace2014}. While some studies go to great lengths to ensure they have properly annotated samples, others base annotation simply on the presence or absence of hashtags or emotes or, even worse, use a third-party algorithm to determine if the Tweet is sarcastic. If annotated using sarcasm markers or self-labelled datasets, it is important to note that the labels may be applied incorrectly, often due to a lack of standardized methods. Examples of this would be improper hashtag labels, either due to how it is used in a Tweet ("\#sarcasm: not just for edgy teens anymore") or poor labelling by the user ("not \#sarcasm"). Emotes similarly have a variety of uses and while ones like :p, ;) and ;P are often associated with sarcasm; there is definitive ambiguity due, once again, to lack of standardization. Third-party tools for annotating are particularly risky as the proposed model is no longer detecting sarcasm but instead detects what the third-party tool annotated as sarcastic. Thus, any error in the third-party sarcasm detection algorithms will carry over.

Issues also occur when researchers annotate tweets and posts. Two crucial factors are author intent and context. From a Tweet in isolation, both can be difficult to determine, especially if the Tweet lacks sufficient context. An example of this would be "My team definitely won't win" if supporting an underdog team in a sporting event. If this statement is made at the start of the game, it would likely be sincere; however, it would be sarcastic if it is made at the end of the match where the team has a significant scoring advantage. Additional issues with researcher labelled data are when domain-specific information is required or when the researchers have a bias in the Tweet's content, as in the example "The president is doing a great job". This can severely compromise the dataset and add unwanted biases, thus bringing the integrity of the results into question.

In closing, automated sarcasm detection is a rapidly growing area of interest in NLP. There is a definitive trend towards using more deep learning approaches and moving away from using conventional machine learning and regression classifiers. This coincides with a trend towards using induced instead of discrete features. Due to the nature of sarcasm, context is essential. Some current detection methods make extensive use of historical data to either provide context or determine if a particular Tweet is out of character for the specific user. Unfortunately, there is a trade-off in efficiency when using historical data for context to improve accuracy.

\section*{Appendix}
\subsection*{Anatomy of a Tweet}
 Twitter posts or Tweets contained the following features:
\begin{itemize}
 \item \textbf{Character Limit- }Initially 140 characters in, 2017 it was doubled to 280, allowing for more context to be added to a Tweet.
 \item \textbf{Hashtags- }A form of self-labelling a post in the format \#<label>. Some examples include \#Vegan, \#weekend and \#sarcasm. Hashtags are often used to give further context to the content of a Tweet. Trending Hashtags occur when a hashtag is used by a large amount of users during a short period of time, this allows for more engagement with the topic the hashtag is referring to.
 \item \textbf{Emoji/Emoticons- }Small cartoon images used to add context to a Tweet. 
 \item \textbf{images- }An image can be added to each Tweet, usually to give context to the Tweet.
 \item \textbf{Usernames- }A unique identifier for each user that allows for them to be followed, reTweeted and replied to. Verified account status was introduced in 2009 for public personalities to prevent impersonation.
 \item \textbf{Mentions- }A mention is a reference to another user by the format of @<username>, @ Twitter for example.
 \item \textbf{ReTweet- }ReTweets allow for the reposting of other users Tweet within a Tweet. This allows for the growth and spread of users Tweets and may result in gaining additional followers, which will increase the user's sphere of influence.
 \item \textbf{Replies- }Replying to a Tweet links the Tweets, allowing for the development of conversation chains.
 \item \textbf{Following- }Following a username allows for the monitoring of that user's activity and allows for individual users' social networks to grow. Every time a user Tweets, all of their followers receive an update.
 \item \textbf{Privacy Options- } Twitter includes a variety of privacy options, including the ability to mute, block, filter and unfollow users. Also, a user can adjust profile visibility to limit who can view the user's Tweets.
\end{itemize}

\section*{Acknowledgment}

This survey was completed with the support of Lakehead University's Centre for Advanced Studies in Engineering and Sciences (CASES). Additionally, special thanks to Abhigya Koirala for his time creating charts and diagrams for use in this work and to Manmeet Kaur Baxi for her Twitter expertise.

\bibliographystyle{unsrt}  
\bibliography{Sarcasm_Detection}  






\end{document}